\documentclass[final]{cvpr}
\usepackage{color}

\usepackage{times}
\usepackage{epsfig}
\usepackage{graphicx}
\usepackage{amsmath}
\usepackage{amssymb}
\usepackage{soul}
\usepackage{booktabs} 
\usepackage{multirow} 
\usepackage{tabularx} 

\usepackage[pagebackref=true,breaklinks=true,colorlinks,bookmarks=false]{hyperref}

\def\cvprPaperID{1720} 
\def\confYear{CVPR 2021}
\usepackage{nopageno}

\begin{document}

\title{Found a Reason for me? Weakly-supervised Grounded Visual Question Answering using Capsules}


\author{Aisha Urooj Khan$^{1,2}$, Hilde Kuehne$^2$, Kevin Duarte$^1$, Chuang Gan$^2$, Niels Lobo$^1$, Mubarak Shah$^1$\\

$^1$ CRCV, University of Central Florida ,  $^2$ MIT-IBM Watson AI Lab\\





}

\maketitle

\begin{abstract}
   The problem of grounding VQA tasks has seen an increased attention in the research community recently, with most attempts usually focusing on solving this task by using pretrained object detectors. However,  pre-trained  object detectors require bounding box annotations for detecting relevant objects in the vocabulary, which may not always be feasible for real-life large-scale applications.
   In this paper, we focus on a more relaxed setting: the grounding of relevant visual entities
   in a weakly supervised manner by training on the VQA task alone. 
   To address this problem, we propose a visual capsule module with a query-based selection mechanism of capsule features, that allows the model to focus on relevant regions based on the textual cues about visual 
   information in the question. 
  We show that integrating the proposed capsule module in existing VQA systems significantly improves their performance on the weakly supervised grounding task.
%
   Overall, we demonstrate the effectiveness of our approach on two state-of-the-art VQA systems, stacked NMN and MAC, on the CLEVR-Answers benchmark, our new evaluation set based on CLEVR scenes with groundtruth bounding boxes for objects that are relevant for the correct answer, as well as on GQA, a real world VQA dataset with compositional questions. 
We show that the systems with the proposed capsule module  consistently outperform the respective baseline systems in terms of answer grounding, while achieving comparable performance on VQA task.\footnote{Code will be available at \url{https://github.com/aurooj/WeakGroundedVQA_Capsules.git}} 

\end{abstract}

\vspace{-5pt}
\section{Introduction}

\noindent VQA systems have now matured to the point where their usage is increasing in real life applications such as answering questions based on radiology images \cite{abacha2019vqa}, helping visually impaired people \cite{gurari2018vizwiz}, and human-robot interactions \cite{qiu2020multi}.
\begin{figure}[t]
\begin{center}
 \includegraphics[width=\linewidth]{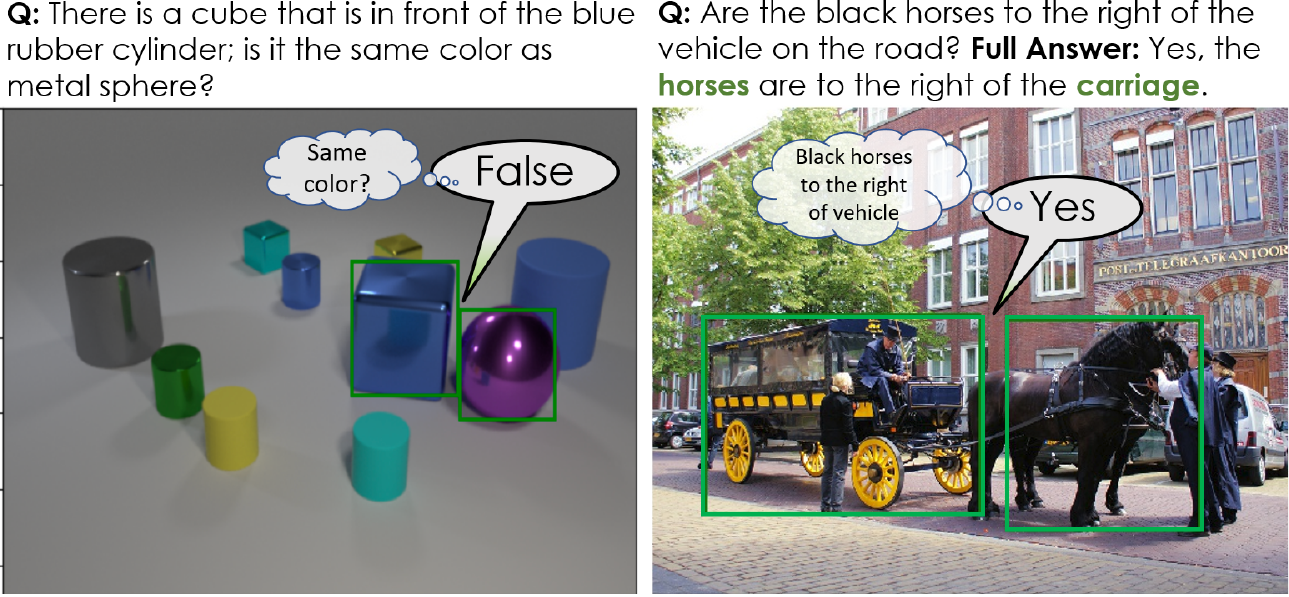}
\end{center}
  \caption{\textbf{Problem definition:} Given an input image and a question, we want to answer the question as well as localize the evidence (shown in green boxes) with VQA supervision alone. Best viewed in color. 
  }
 \vspace{-15pt}
\label{fig:problem-definition}
\end{figure}
However, with the increasing maturity of such systems, it also becomes important to know how the answer is actually generated in order to assess if it is based on the right cues or not. 
If the question is ``Are there black horses to the right of the vehicle?" (see figure \ref{fig:problem-definition}), it may be important to know if the answer is generated because the network found black horses at the right place in the image or not.  
This allows to judge the overall correctness beyond simply evaluating the textual answer. 
Recent works \cite{hudson2019gqa, osman2020towards, chen2020air, Jiang_2020_CVPR}
try to address this problem by starting to evaluate not only the VQA accuracy, but also the accuracy of  grounding that the answer is based on. The grounding of an answer is usually assessed by considering the respective attention map of the image for the given answer, and by evaluating if the objects that are relevant for the right answer are attended to or not. 

To achieve good grounding accuracy, most approaches in this field rely on input feature maps from object detection models that are pretrained with the relevant object classes. This  restricts the scope to known object classes such as MS COCO \cite{lin2014microsoft}, or require to annotate the regions of relevant objects, and to pretrain an object detector for them\cite{hudson2019gqa}. 
Only few attempts have been made so far to address this problem  to train both, the VQA as well as the grounding, without pretrained object detection based on the information of the VQA task alone as e.g. in context of the GQA dataset by only using spatial (appearance)  features \cite{hudson2019gqa}.
This paper focuses on exactly this scenario: weakly supervised 
visual grounding based on VQA supervision. The idea here is that both tasks, the visual question-answering as well as the correct visual grounding, should be learned from the VQA task alone. Hence, we do not use any object-level information as an input or in supervision. 

 
The correct grounding in this case is usually based on two major tasks, finding the relevant visual instances and, usually, modeling the relation between those instances as seen in figure \ref{fig:problem-definition}. 
To address this problem, we propose extending current VQA frameworks with capsules. Capsule networks were introduced by Sabour~\etal \cite{sabour2017dynamic}, and have shown promising results for image interpretability \cite{jung2020icaps} and segmentation in various fields such as 3D point clouds~\cite{Zhao_2019_CVPR}, videos~\cite{duarte2019capsulevos} and medical images~\cite{lalonde2018capsules}. This is the result of capsule layers' ability to learn part-to-whole relationships for object entities through routing-by-agreement. We believe this capability to model objects and their relations qualifies capsules as a good choice for addressing the problem of weakly-supervised grounding in VQA.


Current capsule-based methods follow the practice of adding capsule layers on top of convolutional features, and training them with object class supervision. 
A discrete and supervised masking operation, \ie masking all capsules except the ground-truth class capsule, is often applied to reconstruct or segment the object corresponding to the given class. 
In case of weak VQA grounding, no class or object based supervision is available; only an embedding of a natural language question is given. 
Therefore, we propose a ``soft-masking" procedure which selects the capsule(s) based on the input question. For example, if the reasoning operation is $Find($``blue spheres"$)$, the soft-masking operation will mask all capsules not representing the \textit{``blue spheres"}. 
Once the irrelevant capsules are masked, the capsule representations are passed to future reasoning operations to complete the VQA task.

To evaluate VQA systems for their answer grounding ability, we consider two datasets, the recently proposed GQA dataset \cite{hudson2019gqa} as well as the CLEVR dataset \cite{johnson2017clevr}. To allow the evaluation of grounding accuracy on CLEVR, we propose a new CLEVR validation set, named CLEVR-Answers. CLEVR-Answers provides VQA pairs with the respective ground truth bounding boxes for all objects that the answer is based on. Note that, as we are not interested in using any object annotations during training, {\em we only need ground truth bounding boxes during evaluation, but not during training}. The idea is, thus, to train on the standard CLEVR training set and to learn visual representations of objects during this training without further annotation. We use this new evaluation set to test current state-of-the-art frameworks, MAC~\cite{hudson2018compositional} and Stacked NMN~\cite{hu2018explainable} with respect to their grounding abilities.
We show that, although all frameworks perform at the same level with respect to VQA accuracy, there are major differences with respect to their grounding abilities. 
We show that using capsules with soft query-based masking  significantly improves existing methods' grounding abilities. 




\begin{figure*}
\begin{center}
\includegraphics[width=\linewidth]{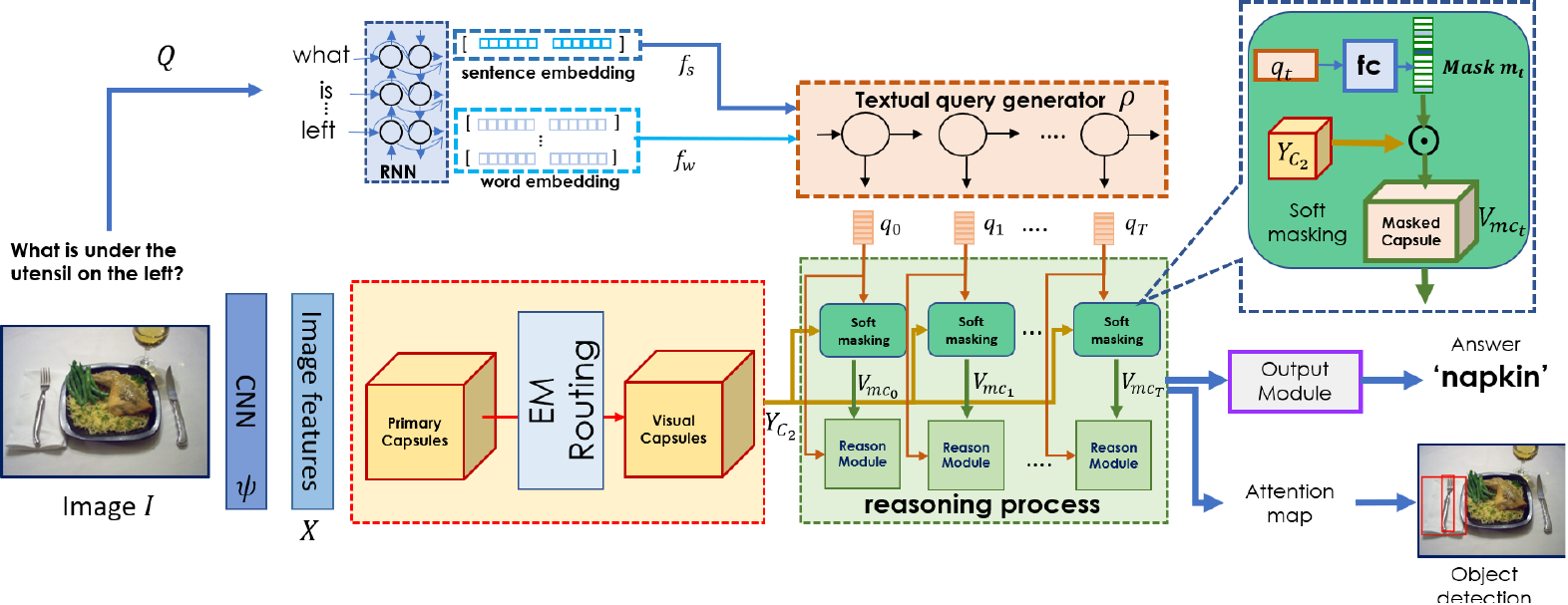}
\end{center}
\vspace{-5pt}
   \caption{Overview of our pipeline: given the question-image pair, we obtain image features $X$ using an image encoder $\psi$ and question features (both sentence $f_s$ and word embeddings $f_w$) using an RNN $\rho$. These question-based features are then input to a multi-hop reasoning module which generates $T$ textual queries $q_0, q_1, ..., q_T$. The Primary Capsules layer then transforms the convolutional image features into capsules (each capsule has a $k\times k$ pose matrix and an activation weight). The primary capsules use a routing-by-agreement algorithm to vote for higher level capsules at each spatial location. These capsules are used as visual representation inside the reasoning process. At each timestep $t \in \{1, 2,..., T\}$, a soft masking module first masks irrelevant capsules using textual query $q_t$ to select a subset of capsules at each spatial location denoted as $V_{mc_t}$. Then, the reasoning operation is performed over selected capsules (using attention combined with other reasoning). Output of the reasoning operation is a vector. Output Module then aggregates these outputs and predicts the answer. We train the system with VQA supervision only. At inference time, we post-process the attention maps produced by the reasoning modules to obtain grounding predictions. }
   \vspace{-15pt}
\label{fig:pipeline}

\end{figure*}

\section{Related Work}
\paragraph{VQA and visual grounding} 
Recent approaches for VQA task rely on object level features as input to improve the VQA accuracy ~ \cite{Anderson2017up-down, lcgn, khan2020mmftbert, mcan, jiang2018pythia, teney2018tips, gan2017vqs, hu2020iterative,8099929}. Those features are extracted from pretrained object detectors. 
This makes the VQA task easier and usually performs better than spatial or appearance features, but it also adds an additional preprocessing step (detecting objects) to the pipeline. Additionally, since the pretraining relies on the object classes in the training set, it limits the extension of such methods to datasets with object-level annotation. Basic appearance or grid-based 
features, \eg, based on a backbone pretrained on ImageNet, are easier to generate and have recently been shown to work as well as object level features~\cite{Jiang_2020_CVPR_Defense} for the VQA task. All these approaches usually only focus on the accuracy of the VQA task, and {\em do not evaluate the respective grounding of their answers}. 

Focusing on this capability, several VQA datasets now provide grounding labels such as GQA \cite{hudson2019gqa}, VCR \cite{zellers2019vcr}, VQS \cite{gan2017vqs}, 	
CLEVRER \cite{yi2019clevrer,chendlc21} and TVQA+ \cite{lei2019tvqa}. Here, object annotations are either provided for all objects in the visual input, or only for the objects relevant to both question and answer.
Out of those, GQA specifically focuses on the evaluation of grounding accuracy with and without object detection supervision and attempts to evaluate MAC \cite{hudson2018compositional} and BottomUp \cite{Anderson2017up-down} for their grounding ability in natural images. We, therefore, choose GQA to evaluate capsule-augmented systems in real world for weakly supervised grounding. Additionally, we compute the answer grounding in terms of overlap and IOU to measure how precise this grounding is in correlation to the answer. 
\vspace{-5pt}
\paragraph{VQA and visual reasoning on CLEVR}
CLEVR \cite{johnson2017clevr} is a diagnostic visual reasoning dataset with compositional questions to test performance of VQA systems on a variety of reasoning skills. Since the introduction of CLEVR, a large number of VQA systems \cite{hu2018explainable, iep, nsvqa, hudson2018compositional, mao2018neuro, Mascharka_2018_CVPR, vedantam2019probabilistic} have surfaced to solve VQA task on this benchmark achieving near perfect VQA accuracy \cite{xnm}. One line of works \cite{iep, hu2018explainable, Mascharka_2018_CVPR} uses reasoning layouts supervision provided with the image-QA pairs. Additionally, neuro-symbolic approaches over object level features are proposed, \eg, in \cite{nsvqa, mao2018neuro}. Many of those ideas implicitly or explicitly include the concept of grounding on this dataset, but usually rely on a pretrained object detector to generate initial object attention maps. 
Several other variants of CLEVR have also emerged trying to solve various problems such as CLEVR-CoGent \cite{johnson2017clevr}, CLEVR-Dialog \cite{kottur2019clevr}, CLOSURE \cite{bahdanau2019closure}, simply-CLEVR \cite{osman2020towards}, and CLEVR-Ref+ \cite{liu2019clevr}, with CLEVR-Ref+ focusing on grounding based on referential expressions, but not VQA, and simply-CLEVR providing only grounding labels for one or all objects in the image. We, on the other hand, provide bounding box labels for all question types without imposing any constraint on the number of objects relevant to answer grounding. Thus, CLEVR-Answers enables us to evaluate grounding abilities of current state-of-the-art methods without any constraints.
\paragraph{Capsule networks}
Hinton \etal \cite{hinton2011transforming} first proposed capsule networks to learn vectors of view equivariant features from images. More recently, Sabour \etal \cite{sabour2017dynamic} extended capsule networks with an iterative routing-by-agreement algorithm to classify and segment multiple digits within an image. Several works have proposed improved methods for routing \cite{hinton2018matrix,zhang2018fast,li2018neural,jeong2019ladder} as well as have applied capsule networks to different tasks and domains \cite{duarte2018videocapsulenet,lalonde2018capsules,Zhao_2019_CVPR,McIntosh_2020_CVPR}. While most previous works tend to be supervised by calculating a loss over a set of ``class capsules", our proposed approach does not have this capsule-to-object supervision; rather, capsules are incorporated in our system as intermediate layers and are learned by using weak supervision from question answers. Several capsule networks which perform classification tasks \cite{sabour2017dynamic,duarte2018videocapsulenet,qin2019detecting} tend to use a masking operation to ensure capsules learn class-specific representations. This operation masks all capsule pose values to 0 except for the selected (i.e. ground-truth or predicted) class capsule, and uses this masked representation to reconstruct or segment the input image or video. Since there are no ground-truth class annotations, we propose a novel soft-masking operation which effectively selects the capsule(s) relevant to the input query and masks irrelevant capsules. 

\section{Proposed Approach}

\subsection{Problem Formulation}
Given an input image $I$ and a question $Q$, our goal is to output the correct answer $a \in A$, where $A$ denotes the answer vocabulary, and $B$ bounding box predictions for the objects which led to the answer $a$. Figure \ref{fig:problem-definition} illustrates the problem. Our pipeline is explained in the following sections. An overview of the framework is given in figure \ref{fig:pipeline}.
\subsection{Input Embeddings}
\noindent \textbf{Question embedding}
We are given a question $Q$ of words $w_1, w_2, ..., w_l$, where $l$ is the length of $Q$ in words. Let $V$ be the vocabulary for question words in the training set with a lookup embedding $E \in \mathbb{R}^{|V|\times d_e}$. Each word in $Q$ is represented by a  $d_e$-dimensional initial embedding vector. Let $\phi(Q, [w_1, w_2, ..., w_l] )$ be a sentence encoder which outputs both sentence level embedding $f_s$ for $Q$ as well as word-level features $f_w$. These sentence-level and word-level embeddings are then input to our system. Following previous works \cite{hu2018explainable, hudson2018compositional}, we choose $\phi$ to be a BiLSTM. Output dimensions for sentence embedding $f_s$ and word embeddings $f_w$ are, therefore, $f_s \in  \mathbb{R}^{d_q}, f_w \in  \mathbb{R}^{d_q}$, where $d_q = 2\times d$, and $d$ is the dimension of sentence encoder. 

\noindent \textbf{Image embedding}
\noindent Given an input image $I$, we compute a feature map $X=\psi(I)$, where $\psi$ is a pretrained image encoder and $X \in \mathbb{R}^{H \times W \times d_f}$ denotes the features extracted for $I$ ($d_f$ is the feature dimension). 

\subsection{Textual query generator}
 \noindent To answer a question based on an image, a VQA system performs attentional parsing of the question, i.e. attends to selected words from the question iteratively depending on the reasoning required to answer the question. This approach of splitting the question into subqueries is often termed as multi-hop or recurrent reasoning, 
 where a query is generated at each reasoning step to attend to the image to collect answer-relevant knowledge. Let $\rho$ be our query generator which takes sentence embedding, $f_s$, and word embeddings, $f_w$, as an input at each time step $t$ ($t = 1,2,..,T$), and outputs query $q_t$ as an output. 
 \begin{equation}
  q_{t} = \rho(f_s, f_w), \forall t \in  \{1,2,...,T\}. 
 \end{equation}
 
 \noindent More details are discussed in the supplementary.

\subsection{Capsules with soft masking}

\noindent A capsule is a group of neurons representing an entity or a part of an entity. In this work, we use matrix capsules \cite{hinton2018matrix}, which are composed of a logistic unit (called the activation) and a 4x4 pose matrix (called the pose). 
The activation indicates the presence of a specific entity, whereas the pose represents the entity's properties. 
A capsule layer consists of many capsules, which use a routing-by-agreement algorithm to vote for capsules in the following layer in order to model part-to-whole relationships. Matrix capsules use EM-Routing algorithm for capsule routing. We integrate them into the process as follows.

\noindent \textbf{Visual capsules:}
From the image embedding, $X$, the primary capsules are obtained by using a learned convolution operation resulting in $C_1$ capsule types each with a 4x4 pose matrix and an activation for each spatial position. The output dimension of the primary capsule layer is $\mathbb{R}^{H \times W \times C_1 \times 4 \times 4}$ and  $\mathbb{R}^{H \times W \times C_1 \times 1}$ for poses and activations respectively. To obtain a higher-level capsule representation, we perform EM-routing over primary capsules to obtain a set of $C_2$ capsules at each spatial position. These capsules model different objects within the scene (including the background). Output dimensions for poses and activations are $\mathbb{R}^{H \times W \times C_2 \times 4 \times 4}$ and  $\mathbb{R}^{H \times W \times C_2 \times 1}$ respectively. They are used as the visual representation of the input image in future steps.



\noindent \textbf{Soft masking:} 
The trivial approach of leveraging this capsule representation for VQA would be to group the poses and activations to form a tensor of shape $\mathbb{R}^{H \times W \times C_2 \times (4\times4+1)}$, and use them as a single feature map like standard convolution-based methods. Although, this performs decently well, it is not an ideal solution since it treats each dimension of the capsule poses as independent features and disregards the fact that all dimensions in the capsule pose represent a single object or entity. 

Instead of this independent feature selection, we propose the selection of individual capsules based on the question. This is achieved by masking capsules which are irrelevant to the reasoning operation. Previous capsule methods use masking for image reconstruction \cite{hinton2018matrix} or segmentation \cite{duarte2019capsulevos}, however they require ground-truth class labels to select the single capsule type which is not masked. Since no object/class-level supervision is present for this task, we propose learning which capsules should be masked in an end-to-end manner. For each reasoning step, a fully connected layer generates a set of $C_2$ logits denoting each capsule types' relevance for the given query. Mathematically, this can be defined as: 
\begin{equation}
  m_{t_{\text{logits}}}=\eta (q_t),
\end{equation}

\noindent where $q_t$ is the textual query at reasoning step $t$, and $\eta$ is the fully connected layer. Then, a one-hot mask, $m_t \in \mathbb{R}^{C_2}$ is generated where $m_i=1$ for $i=argmax(m_{t_{\text{logits}}})$. This mask is then  applied to the visual capsule layer:
\begin{equation}
V_{mc_t} = m_{t} \odot Y_{c_2},
\end{equation}

\noindent where, $Y_{c_2}$ is the output of visual capsules layer, and $V_{mc_t}$ are the masked visual capsules corresponding to textual query $q_t$. We call this operation hard masking.

We find that hard masking operation leads to sub-optimal performance, since the lack of supervision leads to some capsule never being selected, resulting in poor representations. To remedy this, we present a novel soft masking method as visualized in the green box in figure \ref{fig:pipeline}, which allows gradients to flow through all capsules. Instead of creating a one-hot mask, a softmax operation is used on the logits to create a set of soft weights, which then mask the visual capsules, as follows:
\begin{equation}
V_{mc_t} = softmax(\eta(q_t)) \odot Y_{c_2}.     
\end{equation}
These masked visual capsules are then used for reasoning operations as defined by their respective modules. We show that incorporating capsules and soft masking within an attentional VQA system can boost its grounding ability significantly without compromising VQA accuracy, therefore, reducing the performance-explainability trade-off.




\subsection{Output module} 
The  reasoning modules output features which are aggregated over reasoning steps and sent to an output module i.e., a classifier which outputs answer scores. For grounding predictions, we consider the spatial attention maps produced by reasoning modules and post-process them to obtain the object detections. The post-processing is described in the following section.
\begin{figure}[t]
\begin{center}
 \includegraphics[width=\linewidth]{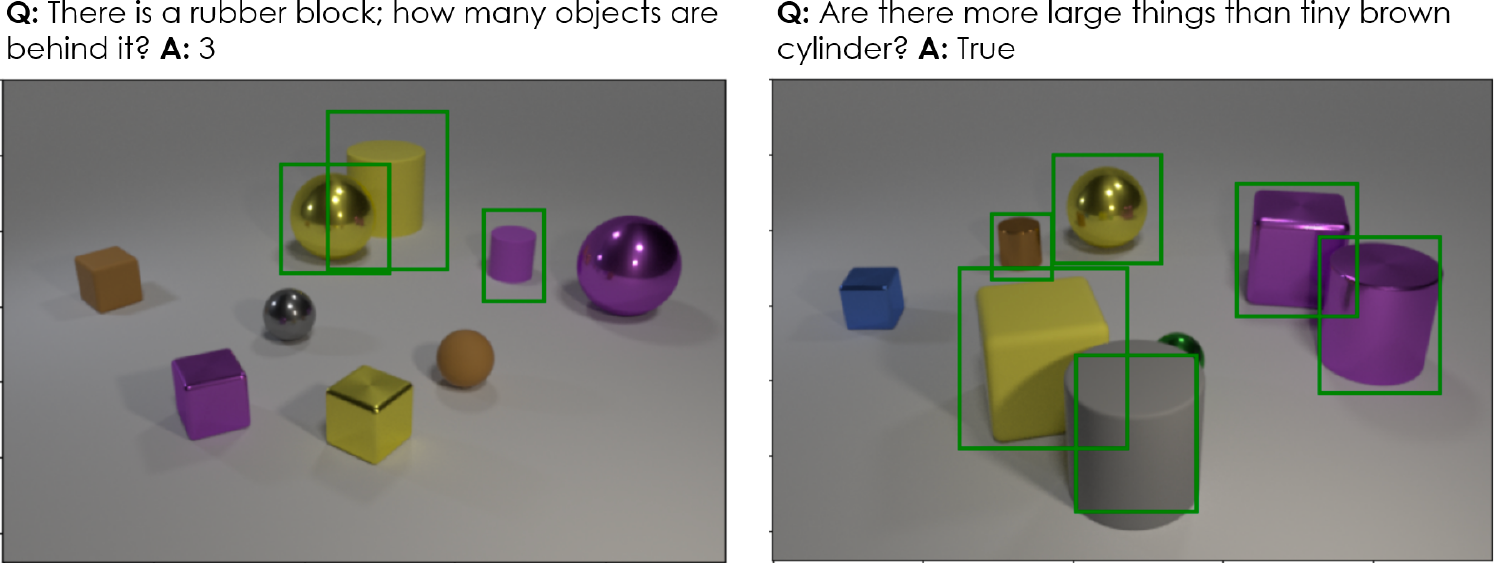}
 \vspace{-15pt}
\end{center}
  \caption{Sample images with QA pair and generated answer grounding labels for CLEVR-Answers dataset. 
  }
 \vspace{-15pt}
\label{fig:data-samples}
\end{figure}
\section{Implementation details}
We integrate capsules into two baseline VQA systems: Stacked Neural Module Networks \cite{hu2018explainable} and MAC \cite{hudson2018compositional}. We make the following architectural changes to these systems.
\vspace{-5pt}
\paragraph{Capsules with MAC. }
MAC \cite{hudson2018compositional} is a recurrent reasoning architecture which performs T reasoning steps to answer the question. Each reasoning step involves generating a question-based control signal (textual query), using this control signal to read from image features (using attention), and writing memory. The final output after T reasoning steps is then combined with the question and goes into the answer classifier. MAC also produces interpretable attention maps for explaining the reasoning process behind VQA. For capsule integration into the MAC cell, we make the following changes: First, capsule layers are added on top of the convolutional layer to obtain visual capsules from image features.
The read module is responsible for attending to spatial image features and retrieving query-relevant image features, based on the previous output and the current control signal (question based feature at timestep t). Inside the read module, we first map the control signal to a feature vector of dimensions  $C_2 \times (4 \times 4 + 1)$ using a trainable linear layer. This feature vector is then used to generate a soft mask to obtain only query-related capsules for further reasoning. Weights for the masking layer are shared among MAC cells. These masked capsules are then used for further reasoning inside the \textit{read} module. 

\vspace{-5pt}
\paragraph{Capsules with Stacked Neural Module Network (SNMN).}
Stacked neural module network is an attentional VQA method following the same reasoning pipeline as explained above. SNMN produces human interpretable attention maps. SNMN trains  convolutional layers on pre-trained image features. The output of these convolutional layers then goes into the reasoning modules and uses textual query to perform the reasoning operation producing an attention map as output. To integrate capsules into SNMN, we append our capsule module on top of image features to obtain $C_2$ visual capsules. Instead of convolutional image features, reasoning modules now perform their reasoning operation on capsules. For query-based soft masking, each neural module has a fully connected layer which takes textual query $q_t \in \mathbb{R}^d$ as input and outputs a feature vector of dimension $C_2 \times (4 \times 4 + 1)$. This feature vector is then used to generate capsule mask of size $C_2$, and for further interaction between query and masked capsules. 
Each reasoning module in SNMN has its own masking layer except $Scene$, $And$, and $Or$, since these modules do not use the textual parameter in their computations.
\vspace{-5pt}
\paragraph{Generation of attention maps. } \label{post-processing} During training, capsule layers learn to attend to different visual cues in the image, including background regions when no grounding evidence is available for the answer. In order to give more weight to high attention regions and suppress attention on the background, we introduce an opacity parameter $\alpha$. For uniform attention regions, opacity is scaled up by $\alpha$.  
After post-processing spatial attention using $\alpha$, an attention threshold of 0.5 is applied to get a binary mask with high attention regions. Each connected component in this binary mask is considered an object detection. See supplementary for results w.r.t. variations in $\alpha$.

\section{Datasets}
We perform our evaluation on two datasets: GQA and CLEVR. 
GQA, as real world dataset for visual reasoning and compositional question answering, combines the two aspects of the proposed idea by providing an evaluation of grounding VQA tasks. It also provides a baseline for the weakly supervised grounding task by using only spatial features that are not pretrained on object annotations. CLEVR, as opposed to GQA, provides a sandbox for visual reasoning VQA tasks with synthetic images only, no visual overlap to any ImageNet categories, and a challenging grounding setting with objects in various combinations of color, shape, size, and material. Recent work on attentional VQA systems (SNMN \cite{hu2018explainable} and MAC \cite{hudson2018compositional}) show high VQA accuracy, making this dataset a good candidate to explore the relationship of grounding and VQA accuracy.  
\vspace{-10pt}
\paragraph{GQA. }
GQA is a real world visual reasoning dataset with multi-hop reasoning questions. GQA provides compositional questions for challenging real world images. Questions in GQA are more diverse than VQA 2.0 \cite{goyal2017making} in several ways with more coverage to relation, spatial, and compositional questions \cite{hudson2019gqa}.
This dataset consists of 22M QA pairs for more than 113K images. GQA provides grounding labels for objects referenced in the question and answer which makes it a suitable test bed for our task. We use the balanced version of this dataset with the standard split provided by the authors for our experiments.  

\begin{table}[t]
\scriptsize
\renewcommand{\arraystretch}{.9}
  \centering \setlength{\tabcolsep}{.5\tabcolsep}   
    \begin{tabular}{lllcccccccc}
       \toprule

    &  & &  & \multicolumn{3}{c}{Overlap} & & \multicolumn{3}{c}{IOU} \\
    \cmidrule{5-7} \cmidrule{9-11} 
   \multicolumn{1}{c}{\textbf{Method}} & \textbf{T} & \textbf{\#param} & Acc.
     &  P & R & F1 & & P & R & F1  \\
       

        \midrule 
     MAC \cite{hudson2018compositional}  & \multirow{2}{*}{4} & {12.20M} & \textbf{97.70}  & 24.92 & 56.27 & 34.55 & & 13.99 & 33.50 & 19.73 \\
     MAC-Caps  &  & {12.92M} & 96.79 & \textbf{47.04} & \textbf{73.06} & \textbf{57.23} & & \textbf{23.97} & \textbf{39.06} & \textbf{29.71}\\
     \\
    MAC \cite{hudson2018compositional}  & \multirow{2}{*}{6} & {12.72M} & 98.00  & 30.10 & 52.41 & 38.24 & & 12.59  & 23.62 & 16.42\\
    MAC-Caps  &  & {12.76M} &\textbf{98.02} & \textbf{48.49} & \textbf{79.75}  & \textbf{60.31} & & \textbf{29.03} & \textbf{47.63} & \textbf{36.07}\\
    \\
    
   MAC \cite{hudson2018compositional}  & \multirow{2}{*}{12} & {14.30M} & \textbf{98.54} & 28.66 & 53.27 & 37.27 & & 8.50 & 18.11 & 11.57\\
    MAC-Caps  &  & {15.02M} & 97.88  & \textbf{50.90} & \textbf{94.61} & \textbf{66.19} & & \textbf{27.72} & \textbf{49.84} & \textbf{35.62}\\
    
     \midrule 
     SNMN \cite{hu2018explainable} & \multirow{2}{*}{9}  & 7.32M & 96.18 & 52.87 & 67.03 & 59.12 & & 37.81 & 47.50 & 42.11\\
     SNMN-Caps & & 6.94M & \textbf{96.66} & \textbf{73.81} & \textbf{78.13} & \textbf{75.91} & & \textbf{50.58} & \textbf{51.80} & \textbf{51.18}\\

	\bottomrule
    \end{tabular}
    \vspace{1pt}
    \caption{\small Comparison with baseline systems on CLEVR-Answers validation set. MAC-Caps and SNMN-Caps are the variants with the proposed soft masked capsules. For MAC, results are shown with varying reasoning steps,  T (column 2). SNMN uses T=9.
    See section \ref{main-results} for details. Numbers are reported in percentages.  }
    \vspace{-15pt}
    \label{tab:main-results}

\end{table} 
\vspace{-10pt}
\paragraph{CLEVR-Answers for Visual Grounding. }
In this paper, we extend CLEVR dataset to CLEVR-Answers for visual grounding of answers. 
The CLEVR dataset is a synthetically rendered dataset for the evaluation of visual reasoning and complex VQA tasks.
It consists of a train set with  70K images and approximately 700K question-answer pairs and a validation set of 15K images with about 150K question-answer pairs. 
To allow for an evaluation of visual grounding on this task, we use the framework provided by \cite{johnson2017clevr},  and generate new question-answer pairs with the bounding box labels for the answers as shown in figure \ref{fig:data-samples}. 
We use the same training and validation scenes (images) and generate 10 new QA pairs for each image. 
To get localization labels for each answer, we follow a two step process: First, we obtain the set of object ids which leads to the answer. Each question in CLEVR dataset is accompanied with a question graph, a stepwise reasoning layout with the information required to solve the question \cite{johnson2017clevr}. We traverse question graph in a backward direction starting from the last node and do breadth-first-search (BFS)  till we traverse all nodes which are at breadth level=1. This gives us the list of objects which were used in the final reasoning step and generated an answer. Please note that not every answer will have grounding labels. For instance, if the question is \textit{``how many blue rubber blocks are behind red cylinder?"} and the answer is 0, then there will be no bounding box labels. Second,  to get bounding boxes for this set of objects, we need scene information. For each question and its corresponding answer grounding objects, we use the center pixel coordinate information (available with each scene object) to locate each object in the scene. Then, based on the object size and shape, we use a few heuristics to get a rough estimate of the bounding box around each object of interest. 

\noindent This two-step process results in 901K bounding boxes (for about 700K QA pairs) for training set and 193K boxes (for about 150K QA pairs) for validation set i.e. more than 1M bounding boxes labels. Note that we do not use those bounding boxes for training, but we will provide them as well to spur further research. 
To have a standard train-val-test setup for our experiments, we separate 1K training images with 10K QA pairs for validation of hyper parameters. The original CLEVR validation set is used as test set and is never seen during training or validation.

\begin{table}[t]
\footnotesize
\renewcommand{\arraystretch}{.9}
  \centering \setlength{\tabcolsep}{.4\tabcolsep}   
    \begin{tabular}{llcccccccc}
       \toprule

    &  & & \multicolumn{3}{c}{Overlap} & & \multicolumn{3}{c}{IOU}   \\
    \cmidrule{4-6} \cmidrule{8-10} 
   \multicolumn{1}{c}{\textbf{Method}} & \textbf{Grd. GT} & Acc. & P & R & F1 & & P & R & F1 \\
      
        \midrule 
                             
    MAC & \multirow{2}{*}{Q}  & \textbf{57.09}  & 19.75 & 30.69 & 24.04 & & 2.88 & 4.36 & 3.46\\
    MAC-Caps &  & 55.13 & \textbf{37.77} & \textbf{63.65} & \textbf{47.41} & & \textbf{5.39} & \textbf{8.65} & \textbf{6.64}\\
    \\
    MAC & \multirow{2}{*}{FA} & \textbf{57.09} & 22.43 & 31.35 & 26.15 & & 3.30 & 4.48 & 3.80\\
    MAC-Caps &  & 55.13 & \textbf{41.53} & \textbf{63.00} & \textbf{50.06} & & \textbf{6.14} & \textbf{8.85} & \textbf{7.25}\\
    \\
   
    MAC & \multirow{2}{*}{A} & \textbf{57.09} & 5.61 & 27.36 & 9.31 & & 0.92 & 4.46 & 1.52\\
    MAC-Caps &  & 55.13 & \textbf{11.95} & \textbf{62.56} & \textbf{20.07} & & \textbf{2.32} & \textbf{11.91} & \textbf{3.88}\\
    
     \midrule  
    MAC & \multirow{2}{*}{All} &  \textbf{57.09} & 25.01 & 30.48 & 27.47 & & 3.66 & 4.28 & 3.95\\
    MAC-Caps &  & 55.13 & \textbf{46.06} & \textbf{62.30} & \textbf{52.96} & & \textbf{7.03} & \textbf{8.72} & \textbf{7.79}\\
   
    \hline
    \end{tabular}
 \vspace{5pt}
    \caption{\small Results on GQA validation set for MAC with T=4. Results are based on grounding of objects referenced in the question (Q), full answer (FA), short answer (A), as well as combined grounding of question and answer (All). We consistently outperform MAC in all metrics. When evaluating for a certain grounding label type, other detected objects are treated as false positives. 
    Numbers are reported in percentages.}
    \label{tab:GQA_grounding_MAC}
\vspace{-15pt}
\end{table} 



      
    


\section{Experiments and Results}

\paragraph{Evaluation Metrics.}
To evaluate the correct answer localization (grounding), we report precision, recall, and F1-score based on two criteria: intersection over detection (Overlap), and intersection over union (IOU). 
Bounding boxes for the object detections are compared with the ground truth bounding boxes to evaluate how close they are to the ground truth labels in terms of overlap and IOU. Predicted regions are considered true positives if the spatial overlap of predicted bounding box and ground truth bounding box is greater than a certain threshold. The detection threshold is 0.5. The baseline systems use a multi-hop reasoning process producing attention maps for each reasoning step. Since the reasoning process is divided into sub-operations resulting in each operation producing a separate attention map, it is possible that evidence for the correct answer was attended at some intermediate step and not necessarily at the last step. To give advantage to the baseline methods, we consider the best attention map in the reasoning process with respect to F1 score. 

\begin{figure*}
\begin{center}
\includegraphics[width=\linewidth]{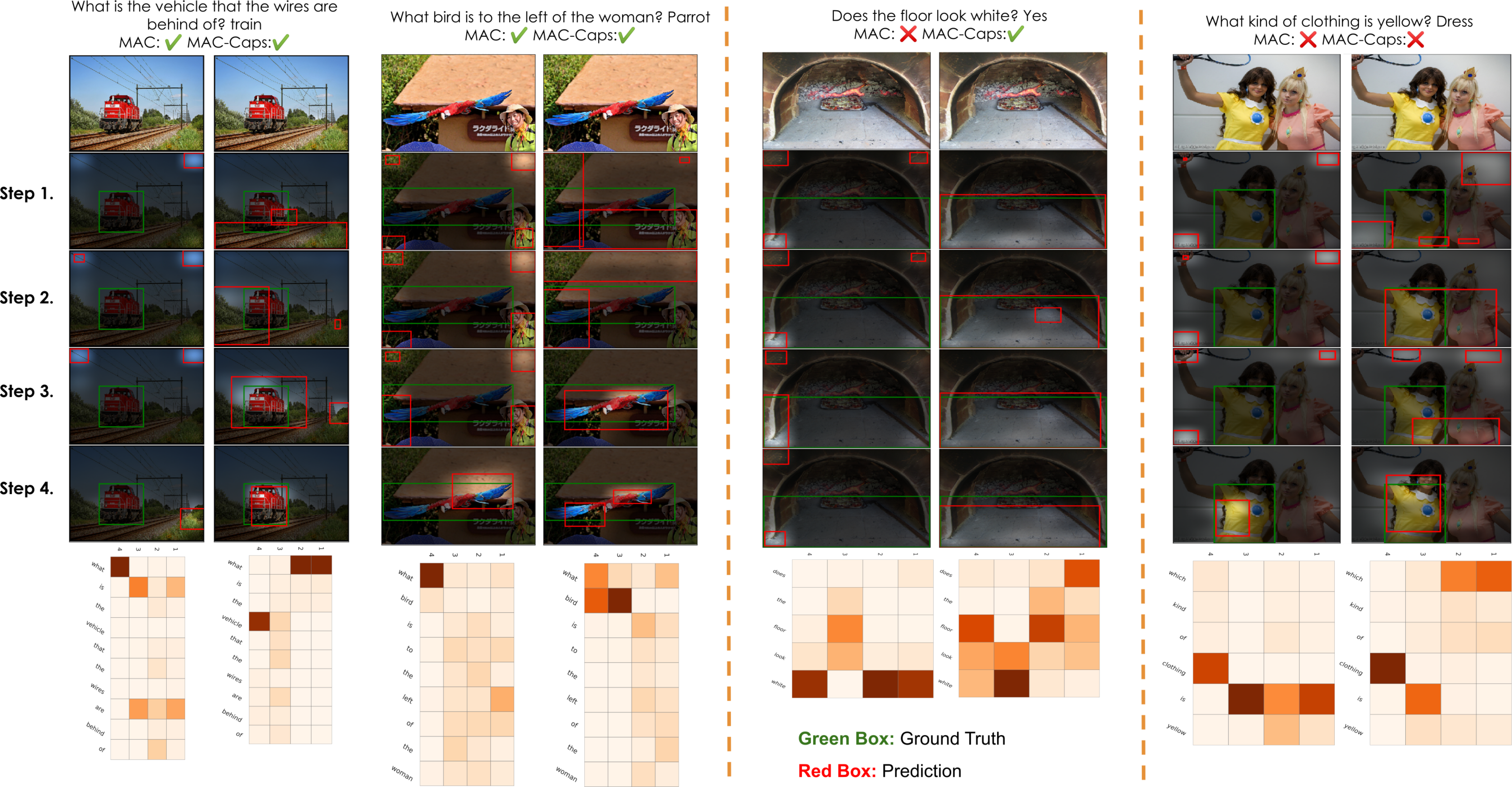}
\end{center}
\vspace{-2pt}
   \caption{\footnotesize Attention visualizations for MAC on GQA dataset. Column 1 shows results for MAC, column 2 shows results for MAC-Caps and the same order is followed onwards. Row 1 shows input image, rows 2-5 are attention visualizations for each reasoning step (T$=$4) with ground truth (green boxes) and detected grounding objects (red boxes), followed by attention on question words for each step. See how MAC-Caps attends to the correct boxes. The attention on question words is also improved. Refer to section \ref{qualitative} for further details and discussion. Best viewed in color. 
   }
   \vspace{-15pt}
\label{fig:qualitative}
\end{figure*}

\subsection{Comparison to baseline method}
\vspace{-2pt}
We first compare the impact of the proposed capsule module on the two baseline systems, MAC and SNMN, on the CLEVR-Answers dataset as well as MAC on GQA. See tables \ref{tab:main-results},
\ref{tab:GQA_grounding_MAC} and \ref{tab:GQA_fulleval_MAC}. 
We use SNMN and MAC as our baselines. These VQA systems take a question with image-based holistic features as input and generate answers with interpretable attention maps. Visual capsules module has the same number of capsules in both layers i.e., we set $C_1=C_2=C$ in all our experiments (C = no. of capsules). 
\vspace{-15pt}
\paragraph{CLEVR-Answers:} \label{main-results} We first evaluate the performance of both systems on the CLEVR-Answer benchmark. We extract  14x14x1024 dimensional features from the conv4 layer of a ResNet-101 backbone pretrained on ImageNet, which are referred to as spatial features in \cite{hudson2019gqa}. We pass them through conv layers by MAC and SNMN to generate 14x14x512 dimensional features. We train the models for 25 epochs, using  the model with best VQA accuracy for grounding evaluation.
The original \textbf{MAC} baseline reports their best VQA accuracy with T=12 system \cite{hudson2018compositional}.
However, it is recommended to use four to six reasoning steps to get interpretable attention maps. Thus, we train both, MAC and MAC-Caps for T=4, 6, and 12 ($\alpha$ is set to 1 for MAC).
Table \ref{tab:main-results} shows the difference of both systems, the MAC and the MAC-Caps with respect to visual grounding. The MAC baseline achieves best IOU F1-score of $19.73$ with T=4 whereas MAC-Caps achieves the best IOU F1-score of $36.07$ (19.41\% $\uparrow$) using T=6 without hurting VQA accuracy. Note that for MAC-Caps, the best Overlap F1-score is reached at T=12, which is an indicator that larger attention maps are produced, which are not rated by the Overlap measurement. Overall, we see a significant and constant increase across all evaluated scores of the proposed MAC-Caps compared to the MAC baseline. 
For the evaluation of \textbf{SNMN} and SNMN-Caps, we train both systems with input features as described above and choose the hyperparameters as mentioned in \cite{hu2018explainable}. Note that SNMN, opposed to MAC, uses an expert layout setting, i.e, question graph layouts are used and learned during training. We get our best results with 24 capsules for SNMN-Caps as further evaluated in section \ref{sec:ablation_study}. Using $\alpha=7$ gives us best grounding results for SNMN. Overall, we see a similar increase in performance as for MAC and MAC-Caps, with an IOU F1-score of $42.11$ for SNMN and an IOU F1-score of $51.18$ for SNMN-Caps.
\begin{table}[t]
\scriptsize
\renewcommand{\arraystretch}{1}
  \centering \setlength{\tabcolsep}{.4\tabcolsep}   
    \begin{tabular}{lccccc}
       \toprule

   \multicolumn{1}{c}{\textbf{Method}} & Validity & Plausibility & Consistency & Distribution & Grounding  \\
      \midrule 
      MAC &  95.14 & 91.34 & \textbf{84.90} & 6.44  & 41.68 (30.34)\\
      MAC-Caps & \textbf{95.17} &  \textbf{91.48} & 80.90 & \textbf{5.67} &  \textbf{45.54 (38.82)}  \\
     
    
    \hline
    \end{tabular}
\vspace{2pt}
    \caption{\small Results on GQA validation set for other evaluation metrics. Grounding results are shown for attention maps from the last (mean) reasoning step(s). Numbers are reported in percentages. }
    \vspace{-7pt}
    \label{tab:GQA_fulleval_MAC}
\end{table} 
 \vspace{-25pt}
\begin{table}[t]
\scriptsize
\renewcommand{\arraystretch}{.9}
  \centering \setlength{\tabcolsep}{.6\tabcolsep}   
    \begin{tabular}{lcccccccc}
       \toprule

      &   & \multicolumn{3}{c}{Overlap} & & \multicolumn{3}{c}{IOU} \\
    \cmidrule{3-5} \cmidrule{7-9} 
   \multicolumn{1}{c}{\textbf{Method}} & Acc.
     &  P & R & F1 & & P & R & F1 \\
      
        \midrule 
     (1)  masked conv.  &  95.69 & 59.71 & 71.13 & 64.92 & & 42.28 & 49.24 & 45.49 \\
     (2)  hard masking & 88.48 & 63.24 & 72.76 & 67.67 & & 43.35 & 48.29 & 45.69\\ 
     (3)  shared mask layer  & 95.76 & 70.40 & 76.13 & 73.15 & & 48.01 & 49.77 & 48.87\\
        \midrule 
     (4)  w/mask (C=8)  & 95.34 & 63.12 & 74.38 & 68.29 & & 42.08 & 47.61 & 44.67 \\
     (5)  w/mask (C=16)   & 95.79 & 73.72 & 76.27 & 74.97 & & \textbf{50.82} & 50.02 & 50.42\\
     (6)  w/mask (C=24)  & \textbf{96.66} & \textbf{73.81} & \textbf{78.13} & \textbf{75.91} & & 50.58 & \textbf{51.80} & \textbf{51.18}\\

    \hline
    \end{tabular}
 \vspace{5pt}
    \caption{\small Ablations over the design choices for the proposed architecture on CLEVR-Answers val set with SNMN as base architecture. Rows 1-3 show the influence of masking (with 16 capsules), where, masked conv.= masking of convolutional layer, hard masking=one hot masking, shared mask layer=weights for masking layer are shared among reasoning modules; w/mask=soft weights are used to mask the capsules (rows 4-6).
    The lower part shows the impact of number of capsules: C = no. of capsules. 
    }
    \vspace{-15pt}
    \label{tab:ablations}

\end{table} 

\paragraph{GQA:} To assess the performance on real world data, we evaluate our system in context of MAC on the GQA dataset. 
GQA provides grounding labels for question, single word answer and sentence-based answer. 
We compare both setups, MAC and MAC-Caps, using the proposed grounding score based on overlap and IOU (see table~\ref{tab:GQA_grounding_MAC}), as well as the metrics proposed by Hudson et al. \cite{hudson2019gqa} where, for each question-image pair, the grounding score is the sum of attention over ground truth region(s) \textit{r}, averaged over all data samples (see table \ref{tab:GQA_fulleval_MAC}). 
We use T=4 for both the MAC baseline and MAC-Caps, showing the best performance on this dataset. We report results on the GQA validation set. 

We again observe that MAC-Caps consistently outperforms the MAC baseline on all metrics in table \ref{tab:GQA_grounding_MAC}. We notice significant improvement (23.91\% $\uparrow$ on F1-score) in terms of Overlap with 3.45\% improvement in F1-score in terms of IOU for full answer grounding. Note that the scores, especially in context of IOU, are much lower on this dataset compared to the CLEVR benchmark, which can be attributed to the complexity of the natural images in this context. Regarding the comparison to the metrics proposed by~\cite{hudson2019gqa} shown in table~\ref{tab:GQA_grounding_MAC}, we see the increase with respect to the grounding abilities of the MAC-Caps compared to the MAC baseline as well as compared to the reported spatial feature baseline of $43\%$ in~\cite{hudson2019gqa}. Overall, both evaluations show that the proposed capsule module allows for a better learning of visual grounding from weak VQA supervision even in a challenging real world setting given with GQA.


\subsection{Ablations and Analysis}
\label{sec:ablation_study}
\noindent \textbf{Convolutional layers vs. Capsules.}
To investigate how much capsules contribute  compared to convolutional layers, we mask convolutional features instead of using capsules. We add a convolutional layer on top of image features resulting in $C \times (4\times4+1)$ features to keep same number of channels as in capsules (here, $C$=16). Similar to soft masking in capsules, these convolutional features are also masked before performing the reasoning operation.  We find that masked convolutional features perform 3.38\% better than the SNMN baseline in terms of IOU, but capsules still outperform them with a large margin (45.49\% vs. 50.42\%) for convolutional masking (see table \ref{tab:ablations} (1)=masked convolutions and (5)=baseline). This shows that query-based masking of capsules performs superior when compared to masked convolutional features .

\noindent \textbf{Hard Masking vs. Soft Masking.}
There are two possible ways to mask capsules based on the query input. The first is masking them using softmax scores which we call soft masking; the second is keeping the capsule with highest probability and mask out the rest of the capsules (using one hot vector as mask), which we call hard masking. We find that using soft masking gives best results. When using hard masking of capsules (C=16), it hurts VQA accuracy (88.07\%), although giving comparable results on grounding metrics (see table \ref{tab:ablations} (2)=hard masking and (5)=baseline). Therefore, we use soft masking for all our experiments.

\noindent \textbf{Shared masking vs. separate masking.}
For SNMN, our final architecture uses a separate masking layer for each reasoning module. We also experiment with using a single masking layer with shared weights for all reasoning modules. While shared masking layer yields good results, we get the best grounding scores using separate masking layer (see table \ref{tab:ablations} (3)=shared mask layer and (5)=baseline).

\noindent \textbf{Performance analysis w.r.t. no. of capsules.}
We finally analyze the system with varying number of capsules. We train the SNMN-Caps model with C=8, 16, and 24. All of them perform superior to the original SNMN in terms of grounding while achieving comparable VQA accuracy. With 24 capsules, SNMN-Caps outperforms the baseline SNMN on both VQA and grounding task (table \ref{tab:ablations} (4-6)).

\vspace{-3pt}
\subsection{Qualitative Results} \label{qualitative}
Figure \ref{fig:qualitative} shows qualitative analysis on GQA dataset with MAC-Caps. For samples, where both systems give the correct answer (columns 1-4), we observe that MAC often attends to corners in the image during the intermediate reasoning steps and attends to the region(s) of interest only at the final stage. For instance, on first sample (columns 1-2), MAC never attended to the correct object yet somehow produces the correct answer.  MAC-Caps, on the other hand, pays attention correctly to relevant regions on earlier stages even for the case where the final answer is incorrect (columns 7-8). Additionally, MAC-Caps produces more precise attention than the baseline system. Attention on question words also seems to be improved for MAC-Caps (last row). Columns 5-6 show the case where, better grounding leads the model to predict the answer correctly. 

\section{Conclusion}
\vspace{-2pt}
This work proposes a novel approach for the weakly supervised grounding of VQA tasks. The proposed capsule-based module can be integrated into current VQA systems. To allow a combination of capsules with VQA based text processing, we proposed a soft masking function that further improves weakly supervised answer grounding. We show by evaluating the system on two challenging datasets, GQA and CLEVR-Answers, the impact of the proposed idea to learn a weakly supervised grounding in VQA tasks. 




\vspace{10pt}
\noindent \textbf{Acknowledgements.} \footnotesize We thank reviewers for their helpful feedback. We also thank Hui Wu for the helpful discussions in the initial phase of this project. Aisha Urooj is supported by the ARO grant W911NF-19-1-0356 and Hilde Kuehne is supported by IARPA via DOI/IBC contract number D17PC00341 for this work. The U.S. Government is authorized to reproduce and distribute reprints for Governmental purposes notwithstanding any copyright annotation thereon. \textbf{Disclaimer:} The views and conclusions contained herein are those of the authors and should not be interpreted as necessarily representing the official policies or endorsements, either expressed or implied, of ARO, IARPA, DOI/IBC, or the U.S. Government.

{\small
\bibliographystyle{ieee_fullname}
\bibliography{cvpr}
}

\newpage
\appendix

\section{Supplementary Material}








\noindent In this supplementary document, we discuss the following:
\begin{enumerate}
\item Structure of Query Generator (section \ref{architecture})
\item Query-focused soft masking of capsules (section \ref{capsule-module})
\item Interpreting attention visualizations (section \ref{how-to-read})
\item Qualitative results (section \ref{qualitative})
\item Further results and analysis (section \ref{further-results})
\begin{itemize}
\item Results for best vs. last reasoning step
\item Results comparison w.r.t. question type
\item Results comparison w.r.t. reasoning type
\item Reduction in parameters 
\item Impact of opacity parameter $\alpha$ on grounding
\end{itemize}

\end{enumerate}
\section{Structure of Query Generator} \label{architecture}
The query generator is a recurrent module proposed by MAC \cite{hudson2018compositional}, also used in SNMN \cite{hu2018explainable}, to obtain question-related reasoning operation $q_t$ at each step t. The recurrent module essentially takes its
output from previous timestep $t-1$ i.e., $q_{t-1}$ along with question features $f_s$ and $f_w$ as sentence and word embeddings respectively, and 
generate attention over question words at current timestep $t$. More specifically, the query generator applies a time-step dependent linear transformation on $f_s$ and combines it with the previous reasoning operation $q_{t-1}$ as follows:
\begin{equation}
    u = W_2 ([W_1^t(f_s)+b_1; q_{t-1}]) + b_2
\end{equation}
where, $W_1^t$ and $W_2$ are $d \times d$ and $d \times 2d$ transformation matrices respectively.  To generate attention over question words $f_w$, following is done:
\begin{equation}
    a_w = softmax(W_3(u \odot f_w)+ b_3)
\end{equation}
where, $W_3$ is a $2d \times d$ matrix and $a_w$ are the attention scores over question words. Finally, $q_t$ is obtained by taking the weighted sum over $l$ question words using $a_w$.
\begin{equation}
    q_t = \sum_{w=1}^{l} a_w \cdot f_w
\end{equation}

\noindent \textbf{Note:} We refer the readers to the works MAC\cite{hudson2018compositional} and SNMN\cite{hu2018explainable} for additional details about these systems.
\section{Query-focused soft masking of capsules} \label{capsule-module}
The proposed query-focused soft masking of capsules is a generic module which can be integrated seamlessly into existing VQA systems. First, the convolutional image features are transformed into the visual capsules as explained in the main paper. Then, a question-based feature called textual-query $q_t$ is used to select relevant capsules by soft masking capsules which are irrelevant to the textual query. For a multihop VQA system with T hops, $q_t$ is the textual-query at timestep $t$, where $t \in \{1,2,...,T\}$. These masked capsules are then input to a reasoning module for further processing. What a reasoning module is, depends on how a VQA system is designed to perform reasoning for answering the question. We show integration of this capsule module into two VQA systems: MAC \cite{hudson2018compositional} and SNMN \cite{hu2018explainable} as shown in the main paper, figure 2 and section 4. Adding the proposed capsule module results in significant improvement in grounding in both baseline methods. 


\begin{figure}[t]
\begin{center}
\includegraphics[width=\linewidth]{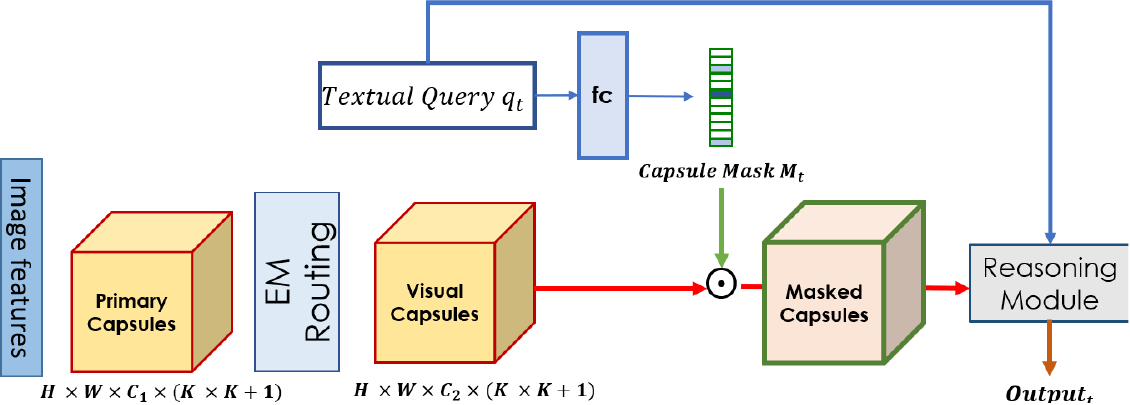}
\end{center}
  \caption{Query-focused Soft Masking of capsules: first, we transform convolutional image features into visual capsules. Then, we use a question-based feature called textual-query $q_{t}, t=1,2,...,T$ to select relevant capsules by soft masking capsules which are irrelevant to the textual query. These masked capsules are then input to a reasoning module for further processing. Reasoning module depends on how a VQA system is designed to perform reasoning for answering the question. We show integration of this module into two VQA systems: MAC \cite{hudson2018compositional} and SNMN \cite{hu2018explainable} as shown in the main paper, figure 2 and section 4. 
  }
\label{fig:capsules}
\end{figure}

\section{Interpreting attention visualization} \label{how-to-read}
\begin{figure}[t]
\begin{center}
\includegraphics[width=\linewidth]{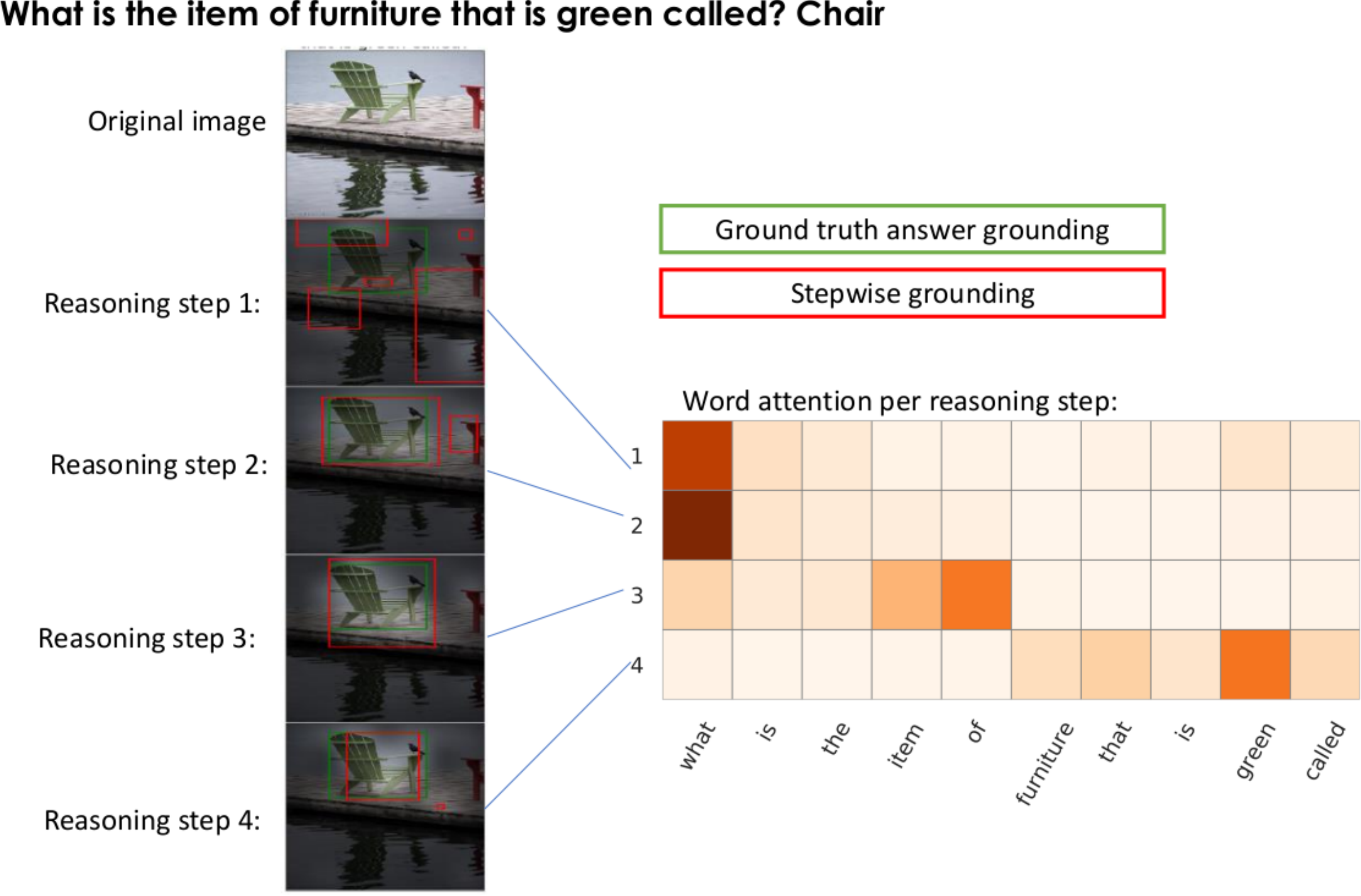}
\end{center}
   \caption{To visualize the attention maps produced at each reasoning step, we  display the original image and question pair, followed by the four reasoning steps, where each attended area is highlighted and marked by a red bounding box. To compare the attended area (grounding) to the ground truth, we also display the ground truth bounding box in green. The attention for each step is guided by the respective word attentions at each reasoning step. 
   }
\label{fig:howtoAttention}
\end{figure}

We start with a short explanation of the qualitative visualization of attention steps. To gain insight into how the VQA system works in combination with grounding, we visualize both, the grounding within the image as well as the respective word attentions within the question. 
As shown in figure \ref{fig:howtoAttention}, we display the original image and the question, followed by the output of the different reasoning steps, \eg, four in case of MAC, with attended regions highlighted and marked by a red bounding box. The ground truth bounding box is shown in green. 

Note that in the paper, we report grounding results for the best overlap between grounding and ground truth, independent of the reasoning step. A non-overlapping bounding box at another reasoning step, \eg, at step one, does not influence the final result. We choose this best-of-all metric, because, especially for SNMN and SNMN-Caps, it is possible that the model might look at the correct grounding at an intermediate step and not necessarily at the last step. See ``Best vs. Last" in section \ref{further-results} for more details.

Additionally, we also show the respective word attentions for each reasoning step. As shown in figure \ref{fig:howtoAttention}, each row shows the attention potential of each reasoning step. Note that for efficient usage of space, this matrix is flipped in all other visualizations and the reasoning steps are shown column wise. 
In the case shown here, one can see that the first two reasoning steps are based on attentions on the word ``What'', the third reasoning step is based on ``item of'', and the last reasoning step attends to ``green''. Note that the term ``furniture''  is not used to answer the question.  

\begin{figure*}[t]
\begin{center}
\includegraphics[width=\linewidth]{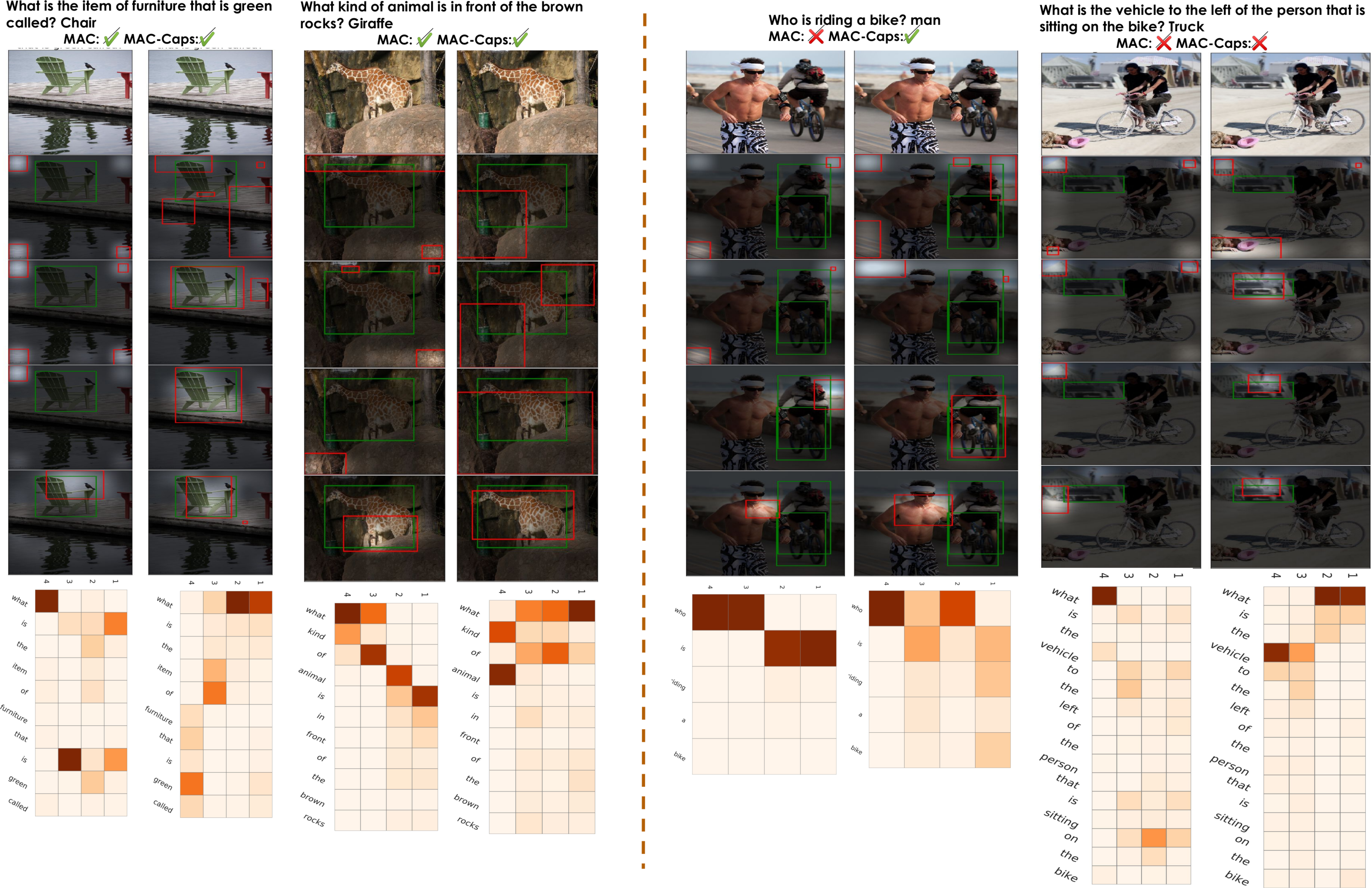}
\end{center}
   \caption{Attention visualizations for MAC on GQA dataset for success (columns 1-4) and failure cases (columns 5-8). Column 1 shows results for MAC, column 2 shows results for MAC-Caps and the same order is followed onwards. Row 1 shows input image, rows 2-5 are attention visualizations for each reasoning step (T=4) with ground truth (green boxes) and detected grounding objects (red boxes), followed by attention on question words for each reasoning step. MAC-Caps attends to the correct boxes with better overlap; for instance,  see example 1, MAC tends to attend corner regions of the image, whereas MAC-Caps starts looking on different image regions and is able to quickly locate ``green'' chair. Similar trend is observed among all examples shown here. The attention on question words is also improved for MAC-Caps with more attention to relevant words when compared to MAC. Refer to section \ref{qualitative} for further details and discussion. Best viewed in color.
   }
\label{fig:qualitative-gqa1}
\end{figure*}


\begin{figure*}[t]
\begin{center}
\includegraphics[width=\linewidth]{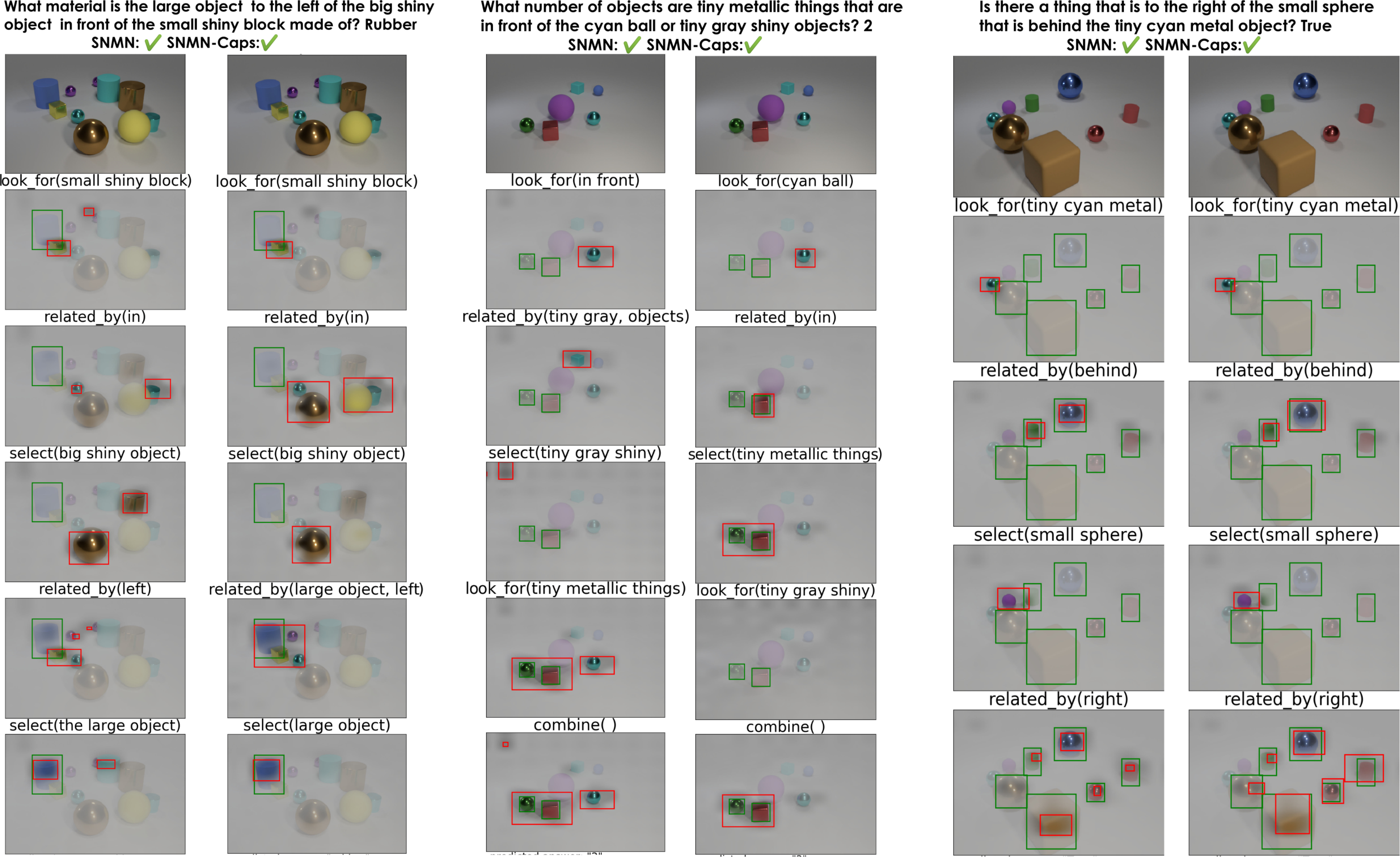}
\end{center}
   \caption{Attention visualizations on the CLEVR-Answers dataset for both SNMN and SNMN-Caps at each reasoning step. We present the question and the input image given to the network in the first row. Each subsequent row is a reasoning step. The reasoning module with the highest weight and the question words with the highest attention (textual query) are displayed above each reasoning step. The red bounding boxes are the detections produced after post-processing the attention maps. The green bounding boxes are the ground-truths, which are displayed for the final reasoning step and the reasoning step in which the respective models achieve the best F1 score. In general, SNMN-Caps produces better groundings, especially at the final reasoning step. 
   }
   
\label{fig:qualitative-snmn1}
\end{figure*}


\begin{figure}[t]
\begin{center}
\includegraphics[width=\linewidth]{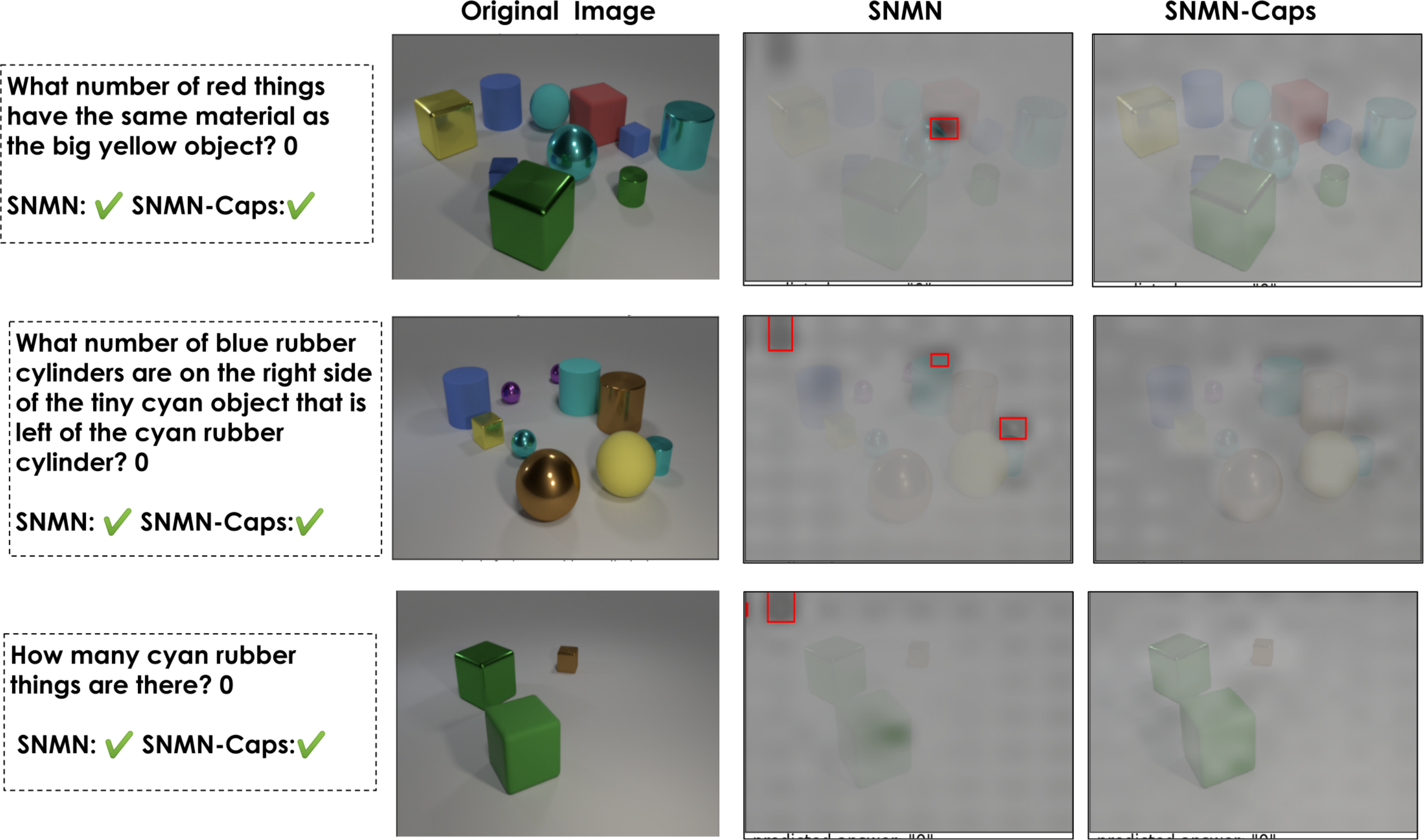}
\end{center}
  \caption{Attention visualizations on the CLEVR-Answers dataset for both SNMN and SNMN-Caps for the final reasoning step on examples with empty grounding maps. Each row displays a data sample with grounding output from SNMN and SNMN-Caps. First two columns show the input image with respective question and answer, third column shows grounding output from SNMN, last column is the grounding from SNMN-Caps. SNMN baseline mostly pays high attentions on regions (red boxes) including objects whereas SNMN-Caps learns to predict nothing by producing uniform attention when the ground truth is supposed to be an empty map.
  }
\label{fig:qualitative-snmn2}
\vspace{-10pt}
\end{figure}


\section{Qualitative Results} \label{qualitative}
\paragraph{GQA} 
Figure \ref{fig:qualitative-gqa1} shows more qualitative comparison between MAC and MAC-Caps. Both MAC and MAC-Caps were trained with T=4 number of reasoning steps. For instance in figure \ref{fig:qualitative-gqa1}'s first example (columns 1-2), MAC attends to question words such as ``is'' at T=1. At T=2, attention is spread, including the word    ``green'' but also paying similar amount of attention to words ``is'' and ``the''. At T=3 and 4, MAC gives most of its attention to words ``is'' and ``what'' respectively. When looking into the question-to-image attention visualization, we observe MAC attending to corners in the image, also in the second step, thus the impact of the attention on green is rather marginal, except for the final reasoning step (last row) where it has an overlap with the correct object based on the attention to ``What''. MAC-Caps, on the other hand, attends to the word ``what'' at T=1,2 with the attention on different regions on image (including ``green chair''). At T=3, MAC-Caps is looking at words ``item of'', and finally attends to the second half of the question at T=4 (with most attention to the word ``green''). MAC-Caps performs much better in localization of ``green chair'' in the image (column 2, rows 4 and 5).  


\paragraph{CLEVR-Answers}
For CLEVR-Answers, we show step-by-step attention visualization for SNMN and SNMN-Caps. Both models have been trained with T=9 reasoning steps, i.e., each question may use up to 9 reasoning steps if that many steps are required to answer the question. For visualization, we follow \cite{hu2018explainable} and remove $No\_Op$ if a question takes fewer than 9 steps to produce an answer. See figure \ref{fig:qualitative-snmn1} for qualitative analysis. We keep the grounding results from the reasoning step giving best grounding output in terms of F1-score. We observe SNMN-Caps is consistent in producing better attention maps covering most of the ground truth boxes. Besides, SNMN-Caps attends to correct objects (or to nothing) based on the input textual query. For instance, in figure \ref{fig:qualitative-snmn1}, column 4, row 5: for the textual query ``tiny gray shiny'', SNMN-Caps produces an empty grounding map, whereas, SNMN attends to a wrong object (cyan cube, column 3, row 4) for the same textual query. Similarly, in the last example (columns 5-6), for textual query ``right'', SNMN misses one object on the right; SNMN-Caps, however, attends to all objects on the right with better overlap to the ground truth grounding boxes. Figure \ref{fig:qualitative-snmn2} shows examples where no grounding evidence was available for the question resulting in an empty map. SNMN fails to produce attention maps with uniform attention in such cases resulting in false positive detections. SNMN-Caps, however,  has learned that it can attend to nothing, thus, correctly generating empty maps in the final step.

\begin{table*}[t]
\scriptsize
\renewcommand{\arraystretch}{.9}
  \centering \setlength{\tabcolsep}{.6\tabcolsep}   
    \begin{tabular}{llcccccccccccccccccc}
       \toprule
        \multicolumn{19}{c}{\textbf{CLEVR-Answers}}  \\
        \midrule
    &   &  & \multicolumn{8}{c}{Overlap} & & \multicolumn{8}{c}{IOU} \\
    \cmidrule{4-11} \cmidrule{13-20} 
    & & &   \multicolumn{2}{c}{P}  & & \multicolumn{2}{c}{R} & & \multicolumn{2}{c}{F1} & & \multicolumn{2}{c}{P} & & \multicolumn{2}{c}{R} & & \multicolumn{2}{c}{F1}  \\
     \cmidrule{4-5} \cmidrule{7-8} \cmidrule{10-11} \cmidrule{13-14} \cmidrule{16-17} \cmidrule{19-20}
   \multicolumn{1}{c}{\textbf{Method}} & \textbf{T} & Acc.
     &  best & last & & best & last & & best & last & &  best & last & & best & last & & best & last  \\
        \midrule 
     MAC \cite{hudson2018compositional}  & \multirow{2}{*}{4} & \textbf{97.70}   & 24.92 & 11.64 & & 56.27 & 29.63 & & 34.55 & 16.72 & & 13.99 & 6.45 & & 33.50 & 16.40 & & 19.73 & 9.25   \\
     MAC-Caps  &  & 96.79 & \textbf{47.04} & \textbf{31.22} & & \textbf{73.06} & \textbf{57.03} & & \textbf{57.23} & \textbf{40.35} & & \textbf{23.97} & \textbf{12.24} & & \textbf{39.06} & \textbf{22.15} & & \textbf{29.71} & \textbf{15.77}\\
     \\
    MAC \cite{hudson2018compositional}  & \multirow{2}{*}{6} & 98.00  & 30.10 & 12.98 & & 52.14 & 25.77 &   & 38.24 & 17.26 & & 12.59 & 5.06 &   & 23.62 & 10.03 &  & 16.42 & 6.73\\
    MAC-Caps  &  & \textbf{98.02} & \textbf{48.49} & \textbf{28.12} & & \textbf{79.75} & \textbf{54.74}  & & \textbf{60.31} & \textbf{37.15} & & \textbf{29.03} & \textbf{13.23} & & \textbf{47.63} & \textbf{25.47} & & \textbf{36.07} & \textbf{17.41}\\
    \\
    
   MAC \cite{hudson2018compositional}  & \multirow{2}{*}{12} & \textbf{98.54} & 28.66  & 9.47 & & 53.27 & 21.33 & & 37.27 & 13.11 & & 8.50 & 2.62 & & 18.11 & 5.89 & & 11.57 & 3.62\\
    MAC-Caps  &  & 97.88  & \textbf{50.90} & \textbf{26.22} & & \textbf{94.61} & \textbf{60.26} & & \textbf{66.19} & \textbf{36.54} & & \textbf{27.72} & \textbf{9.27} & & \textbf{49.84} & \textbf{20.97} & & \textbf{35.62} & \textbf{12.86}\\
    
     \midrule 
        & \textbf{\#param} & & & & & & & & \\
        & $---$ &\\
     SNMN \cite{hu2018explainable}  & 7.32M & 96.18 & 52.87 & 44.13 & & 67.03 & 56.64 & & 59.12 & 49.61 & & 37.81 & 29.51 & & 47.50 &37.32 & & 42.11 & 32.96 \\
     SNMN-Caps  & 6.94M & \textbf{96.66} & \textbf{73.81} & \textbf{63.25} & & \textbf{78.13} & \textbf{67.64} & & \textbf{75.91} &\textbf{65.37} & & \textbf{50.58} & \textbf{40.54} & & \textbf{51.80} &\textbf{41.96} & & \textbf{51.18} &\textbf{41.24}\\

	\bottomrule
    \end{tabular}
    \vspace{5pt}
    \caption{\small \textbf{Best vs. Last:} Comparison with baseline systems on CLEVR-Answers validation set for grounding from best vs. last reasoning step. Best: the reasoning step in which the respective models achieving the best F1-score, last: final reasoning step is used to evaluate for grounding. MAC-Caps and SNMN-Caps are the variants with the proposed soft masked capsules. For MAC, results are shown with varying reasoning steps. SNMN uses T=9. See section \ref{best-vs-last} for details. Numbers are reported in percentages. 
    }
    \label{tab:main-results-clevr-best-vs-last}

\end{table*} 

\begin{table*}[t]
\footnotesize
\renewcommand{\arraystretch}{.9}
  \centering \setlength{\tabcolsep}{.6\tabcolsep}   
    \begin{tabular}{llccccccccccccccccc}
       \toprule
        \multicolumn{19}{c}{\textbf{GQA}}  \\
        \midrule

     &    & \multicolumn{8}{c}{Overlap} & & \multicolumn{8}{c}{IOU} \\
    \cmidrule{3-10} \cmidrule{12-19} 
    & &    \multicolumn{2}{c}{P}  & & \multicolumn{2}{c}{R} & & \multicolumn{2}{c}{F1} & & \multicolumn{2}{c}{P} & & \multicolumn{2}{c}{R} & & \multicolumn{2}{c}{F1}  \\
     \cmidrule{3-4} \cmidrule{6-7} \cmidrule{9-10} \cmidrule{12-13} \cmidrule{15-16} \cmidrule{18-19}
   \multicolumn{1}{c}{\textbf{Method}} & \textbf{Grd. GT} &  best & last & & best & last & & best & last & &  best & last & & best & last & & best & last  \\
      
        \midrule 
                             
    MAC & \multirow{2}{*}{Q}    & 19.75 & 10.79 && 30.69 & 16.38 && 24.04 & 13.01 && 2.88 & 1.39 && 4.36 & 2.09 && 3.46 & 1.67\\
    MAC-Caps &  & \textbf{37.77} & \textbf{17.39} && \textbf{63.65} & \textbf{28.10} && \textbf{47.41} & \textbf{21.49} && \textbf{5.39} & \textbf{1.87} && \textbf{8.65} & \textbf{2.96} && \textbf{6.64} & \textbf{2.29}\\
    \\
    MAC & \multirow{2}{*}{FA}  & 22.43 & 13.62 && 31.35 & 18.63 && 26.15 & 15.74 && 3.30 & 1.80 && 4.48 & 2.42 && 3.80 & 2.06\\
    MAC-Caps &  & \textbf{41.53} &\textbf{19.69} && \textbf{63.00} & \textbf{28.58} && \textbf{50.06} & \textbf{23.31} && \textbf{6.14} & \textbf{2.27} && \textbf{8.85} & \textbf{3.23} && \textbf{7.25} & \textbf{2.67}\\
    \\
   
    MAC & \multirow{2}{*}{A} & 5.61 & 5.05 && 27.36 & 24.44 && 9.31 & 8.37 && 0.92 & 0.76 && 4.46 & 3.70 && 1.52 & 1.27\\
    MAC-Caps &  & \textbf{11.95} & \textbf{5.46} && \textbf{62.56} & \textbf{27.90} && \textbf{20.07} & \textbf{9.13} && \textbf{2.32} & \textbf{0.97} && \textbf{11.91} & \textbf{4.94} && \textbf{3.88} & \textbf{1.62}\\
    
     \midrule  
    MAC & \multirow{2}{*}{All}    & 25.01 & 15.23 && 30.48 & 18.03 && 27.47 & 16.51 && 3.66 & 1.97 && 4.28 & 2.28 && 3.95 & 2.11\\
    MAC-Caps && \textbf{46.06} & \textbf{22.16} && \textbf{62.30} & \textbf{27.98} && \textbf{52.96} & \textbf{24.73} && \textbf{7.03} & \textbf{2.53} && \textbf{8.72} & \textbf{3.10} && \textbf{7.79} & \textbf{2.79}\\
   
    \hline
    \end{tabular}
    \vspace{10pt}
    \caption{\small  \textbf{Best vs. Last:} Results on GQA validation set for MAC with T=4 for grounding from best vs. last reasoning step. Best: the reasoning step in which the respective models achieving the best F1-score, last: final reasoning step is used to evaluate for grounding. Results are based on grounding of objects referenced in the question (Q), full answer (FA), short answer (A), as well as combined grounding of question and answer (All). We consistently outperform MAC in all metrics. When evaluating for a certain grounding label type, other detected objects are treated as false positives. VQA accuracy is reported in the main paper (table 3).
    Numbers are reported in percentages. 
    }
    \label{tab:GQA_grounding_MAC-best-vs-last}
\end{table*} 
\vspace{20pt}

\begin{table}[t]
\footnotesize
\renewcommand{\arraystretch}{.9}
  \centering \setlength{\tabcolsep}{.6\tabcolsep}   
    \begin{tabular}{llcccccc}
      \toprule

      & & \multicolumn{3}{c}{Overlap} & \multicolumn{3}{c}{IOU}   \\
    \cmidrule{3-8} 
  \multicolumn{1}{c}{\textbf{Reason type}} & \textbf{Method}
     &  P & R & F1 & P & R & F1 \\
      
        \midrule 
    \multirow{2}{*}{Count}  & SNMN & 47.28 & 75.43 & 58.13 & 29.43 & 45.65 & 35.79\\
                            & SNMN-Caps & 73.99  & 87.69 & 80.26 & 42.35 & 45.72 & 43.97 \\
                            \midrule
    \multirow{2}{*}{Exist} & SNMN & 26.91 & 74.15 & 39.49 & 16.90 & 46.24 & 24.75\\
                          & SNMN-Caps & 54.83 & 87.70 & 67.48 & 29.56 & 45.49 & 35.83  \\
     \midrule
    \multirow{2}{*}{Comp. Num.} & SNMN & 46.27 & 64.95 & 54.04 & 30.82 & 44.44 & 36.40\\
                                & SNMN-Caps &  58.51 & 78.23 & 66.95 & 35.38 & 48.74 & 41.00  \\
    \midrule
    \multirow{2}{*}{Comp. Attr.} & SNMN & 84.57 & 47.64 & 60.95 & 73.57 & 40.97 & 52.63\\
                                    & SNMN-Caps & 92.62 & 55.08 & 69.08 & 82.85 & 45.40 & 58.66 \\
                                    \midrule
    \multirow{2}{*}{Query Attr. }  & SNMN & 62.93 & 75.39 & 68.60 & 46.94 & 56.69 & 51.36\\
                                & SNMN-Caps & 78.61 & 89.42 & 83.67 & 57.47 & 66.81 & 61.79  \\
                                \midrule
    \multirow{2}{*}{Overall}  & SNMN & 52.10 & 66.48 & 58.42 & 37.38 & 47.38 & 41.79\\
                             & SNMN-Caps & 73.81 & 78.13 & 75.91 & 50.58  & 51.80 & 51.18  \\
    \hline
    \end{tabular}
    \vspace{5pt}
    \caption{\small Results comparison w.r.t reasoning type on CLEVR validation set (for best reasoning step). Numbers are reported in percentages. 
    }
    \label{tab:question-family-results}
\vspace{-10pt}
\end{table} 

\begin{table}[t]
\footnotesize
\renewcommand{\arraystretch}{.9}
  \centering \setlength{\tabcolsep}{.6\tabcolsep}   
    \begin{tabular}{llcccccc}
       \toprule

      & & \multicolumn{3}{c}{Overlap} & \multicolumn{3}{c}{IOU}  \\
    \cmidrule{3-8} 
   \multicolumn{1}{c}{\textbf{Reasoning type}} & \textbf{Method}
     &  P & R & F1 & P & R & F1  \\
      
        \midrule 
    Zero hop  &  & 76.39  & 84.03 & 80.03 & 55.23 & 59.53 & 57.30 \\
    One hop  &  & 69.15  & 86.89 & 77.02 & 47.18 & 58.56  & 52.26  \\
    Two hop  &  &  73.10 & 90.21 & 80.76 & 51.25 & 63.95 & 56.90 \\
    Three hop  &  & 74.59 & 92.02 & 82.39 & 53.36 & 67.81 & 59.72   \\
    Single OR   &  & 88.31 & 88.33 & 88.32 & 59.44 & 50.24 & 54.46 \\
    Single AND  &   & 69.80 & 88.11 & 77.89 & 46.97 & 61.03 & 53.09  \\
    Same relate  &   & 66.76 & 88.78 & 76.21 & 39.39 & 51.02 & 44.46  \\
    Comparison  &   & 92.62 & 55.08 & 69.08 & 82.85 & 45.40 & 58.66  \\
    Compare integer  &   & 58.51 & 78.23 & 66.95 & 35.38 & 48.74 & 41.00  \\
 
    Overall  &   & 73.81 & 78.13  & 75.91 & 50.58 & 51.80 & 51.18 \\
    \hline
    \end{tabular}
    \vspace{5pt}
    \caption{\small Results breakdown w.r.t reasoning type on CLEVR dataset for SNMN-Caps (for best reasoning step). Questions with reasoning types $same\_relate$ and $compare\_integer$ are more challenging (IOU F1-score is $<$ 45\%) for answer grounding than other reasoning types. See section \ref{further-results}, paragraph 3 for more analysis. Numbers are reported in percentages. 
    }
    \label{tab:reasoning-type-results}
\vspace{-10pt}
\end{table} 
\begin{figure}[t]
\begin{center}
\includegraphics[width=\linewidth]{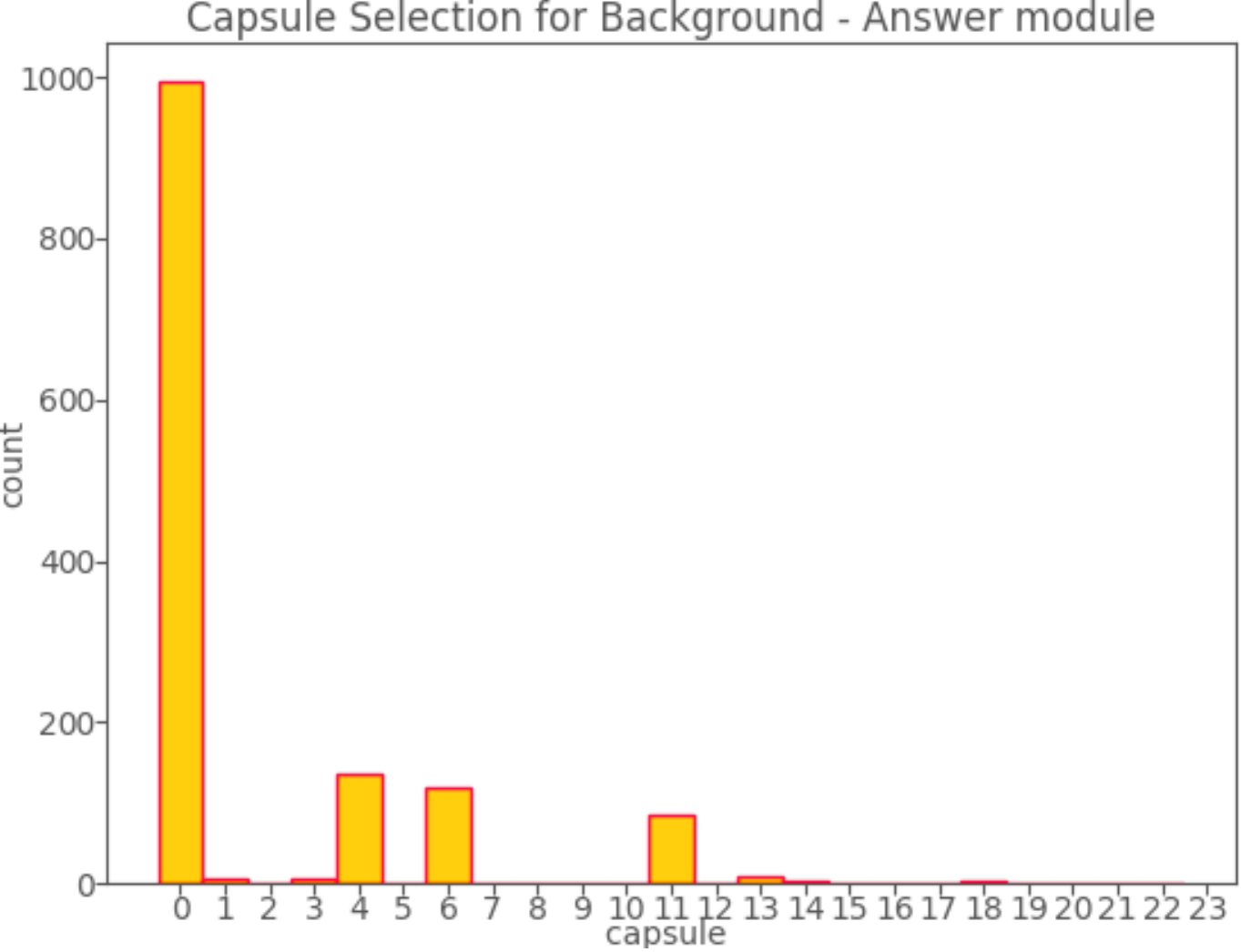}
\end{center}
\vspace{-5pt}
   \caption{Background capsules selected by the $Answer$ module in SNMN-Caps on CLEVR-Answers train-val set for questions with empty grounding maps. MAC-Caps was trained with 24 capsules. X-axis shows the capsule number, and Y-axis shows the frequency (count of questions) of a particular capsule being selected. For each question in this subset, we select the highest probability capsule in the answer module (probability scores are generated with a  soft-masking layer, where, capsules with less probability scores are considered masked or not selected). As we can see that capsule 0 is contributing the most for questions with empty grounding maps. See \ref{background-capsules} for further details. 
   }
\label{fig:capsules-background}
\end{figure}

\begin{figure}[t]
\begin{center}
\includegraphics[width=\linewidth]{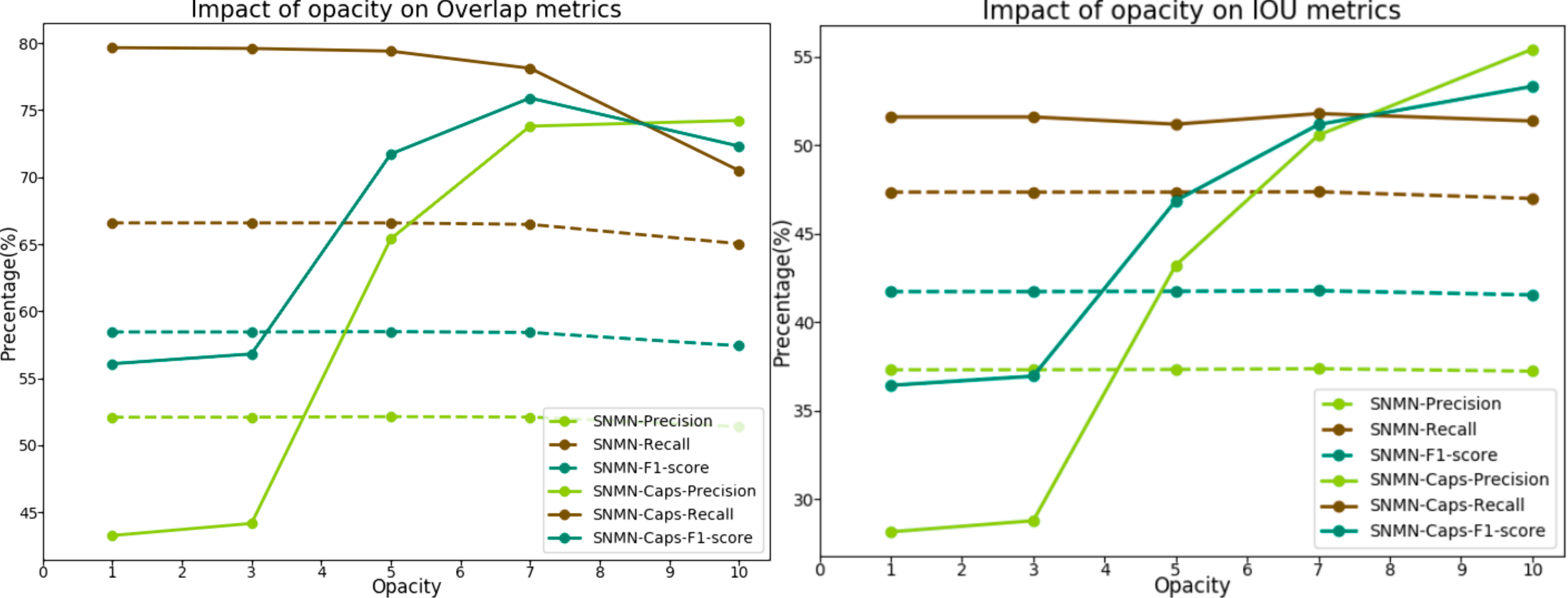}
\end{center}
   \caption{ Impact of opacity parameter $\alpha$ on grounding results for SNMN and SNMN-Caps. Left: impact of opacity parameter $\alpha$ on overlap in terms of precision, recall, and F1-score; Right: impact of $\alpha$ on IOU in terms of precision, recall, and F1-score. Dotted lines are results for SNMN, solid lines display results for SNMN-Caps. SNMN-Caps has significantly higher recall for all values of $\alpha$. However, scaling up opacity on uniform attention regions by $\alpha$ improves precision and consequently F1-score for SNMN-Caps. Results for SNMN are not effected by changing $\alpha$. Therefore, we choose $\alpha=7$ to post process attention maps from both SNMN and SNMN-Caps to report final results. 
   }
   \vspace{-10pt}
\label{fig:opacity-param}
\end{figure}

\section{Further results and analysis} \label{further-results}

\noindent \textbf{Best vs. Last} \label{best-vs-last}
We first compare the baselines and their respective capsule-based variants for best-of-all metric compared to the performance of the grounding of the last reasoning step (in table \ref{tab:main-results-clevr-best-vs-last} and table \ref{tab:GQA_grounding_MAC-best-vs-last} ). We find that scores degrade significantly for the final reasoning step compared to keeping the best score, because a model may attend to the correct answer regions at any step (including intermediate steps) and process them further to produce an answer. 
Nonetheless, we see similar gain in grounding score over baselines even when results from the last step are considered. For MAC on \textbf{CLEVR-Answers}, we observe the gap is reduced between best results for the baseline vs. last results for SNMN-Caps. For recall, SNMN-Caps (last) is always better than SNMN (best). For MAC=4, we observe that SNMN-Caps (last) is better than the best score for SNMN on all metrics in terms of overlap. These observations indicate that adding capsules has improved grounding both for intermediate reasoning steps as well as the final step. We report results for the best grounding step in the main paper. 
For \textbf{GQA}, we evaluate for grounding of relevant objects in the question (Q), sentence-based full answer (FA), single word answer (A), and for all objects in question-answer pair (All). For all grounding label types, we notice a similar drop in grounding scores if the attention map from the final reasoning step is considered for both MAC and MAC-Caps. We still outperform the MAC baseline model in terms of overlap and IOU; however, the gap between F1-scores for IOU is reduced compared to keeping the best score. This is not surprising because when visualizing the attentions produced by MAC, we notice that it usually looks at the correct answer grounding regions in the last step. MAC-Caps, on the other hand, performs better regardless of which reasoning step (best or last) is used for evaluation. 

\noindent \textbf{Results comparison w.r.t. question type.} Table \ref{tab:question-family-results} shows comparison of SNMN and SNMN-Caps models on CLEVR-Answers dataset for different question types, e.g., count, exist, and so on. Although we observe a boost in grounding scores with the proposed capsules module on all question families; we notice that question type $exist$ and $compare\_number$ are the most challenging question types. When looking at the IOU scores, SNMN and SNMN-Caps yield F1-scores of 24.75 vs. 35.83 and 36.40 vs. 41.00, respectively. Both of these question families have boolean (yes/no or true/false) answers, i.e., chance of failure is 50\%. For question type $exist$, the lower grounding performance can be attributed to the boolean nature of these questions. For $compare\_number$ question type, the reasoning operation (hence attention) is split among multiple reasoning steps which also leads to a lower grounding score. In terms of overlap, $count$ and $query\_attribute$ seems to be easier questions for grounding where we observe F1-scores of 80.26 and 83.67 respectively. Overall, with our approach, we obtain 17.49\% and 9.39\% improvement in F1-scores for overlap and IOU respectively.

\noindent \textbf{Results comparison w.r.t. reasoning type.} Table \ref{tab:reasoning-type-results} shows results breakdown of SNMN-Caps on CLEVR-Answers dataset w.r.t. reasoning type--a fine-grained breakdown of grounding results. CLEVR has compositional questions which may need a varying number of reasoning operations to answer them, e.g., a ``two hop" question requires two reasoning hops to be answers. We observe the lowest grounding F1-scores obtained on $compare\_integer$ both in terms of overlap and IOU. This is consistent with the previous observation that question type of $compare\_number$ ($compare\_integer$ in reasoning type) is more challenging for grounding relative to the grounding of other reasoning operations.

\noindent \textbf{Reduction in Parameters.} \label{param-reduction}
Since the capsule representation is more compact than the original image features (16 capsules require $d=16\times16+16=272$ dimensional vector representation, as opposed to the $d=512$ dimensional feature maps generated from convolutions in the baseline systems), operations within the reasoning modules require fewer parameters. When extending SNMN with 16 capsules, the number of learned parameters reduces by 15.67\% (from 7.32M to 6.2M parameters); in MAC with T=4, there is a 7.86\% reduction (17.66M to 16.28M). Even with 16 capsules, capsules perform really well in the grounding task (see table 4 in the main paper) indicating that capsules inherently have an advantage over its convolutional variants even with fewer parameters. For grounding, we see similar performance in MAC with 16 capsules using less parameters. However, for MAC, we use C=32 for network length T=4, C=24 for T=6, and C=32 for T=12 because of the best scores on VQA task. 

\noindent \textbf{Impact of opacity parameter $\alpha$ on grounding.}
To obtain grounding detections from the attention maps, we introduce an opacity parameter $\alpha$. Specifically, \cite{hu2018explainable} used $\alpha$=3 to suppress uniform attention regions by upscaling of opacity in those regions. For SNMN-Caps, we observed some capsules were activated on the background, particularly when no object of interest is found in the image. Although, we find that SNMN-Caps has high recall when compared to SNMN, increasing $\alpha$ improved precision of attention maps which led to the increased F1-score for both overlap and IOU. We perform same post processing on SNMN and SNMN-Caps to report numbers. We noticed the scores for SNMN remain unaffected by parameter $\alpha$ unless increased to a very high value. Figure \ref{fig:opacity-param} shows impact of opacity on grounding results.

\noindent \textbf{Capsules can model background.} \label{background-capsules}
While studying the capsules' behavior, we observe that our model has an advantage on samples with no ground truth boxes. More specifically, we take SNMN-Caps model trained with CLEVR-Answers and used train-val split for this study. Interestingly, we find that some capsules are focusing more on the background. When carefully examined for
examples where the grounding output should be an empty map, we find that capsules are looking at the background for 677 out of these 1586 samples rather than focusing on any object, and are better than the baseline for 83.17\% of such cases. 
The original SNMN model has clearly not learned this behavior and always focuses on some object in the last reasoning step which leads to false positive detections. To further investigate this subset of questions, we look into the capsules with highest probability for the last step before $No\_Op$ (no operation); we notice that capsule 0 was selected the most for the $Answer$ module (see figure \ref{fig:capsules-background}).
This validates our observation that capsules have learned to attend the background when no evidence is available for the answer.
See figure \ref{fig:qualitative-snmn2} for attention visualizations with and without the capsule module.

\noindent Code and the CLEVR-Answers dataset will be released upon publication. 

\end{document}